\documentclass{article}

\usepackage{microtype}
\usepackage{graphicx}
\usepackage{subcaption}
\usepackage{booktabs} 

\usepackage{hyperref}

\usepackage{algorithmic}
\usepackage[ruled, vlined, linesnumbered]{algorithm2e}

\usepackage[preprint]{icml2026}

\usepackage{amsmath}
\usepackage{amssymb}
\usepackage{mathtools}
\usepackage{amsthm}

\usepackage{bbding}
\usepackage{diagbox}
\usepackage{subcaption}

\usepackage{cancel}

\usepackage{multirow}
\usepackage{booktabs}       
\usepackage{rotating}

\usepackage[capitalize,noabbrev]{cleveref}

\theoremstyle{plain}

\theoremstyle{definition}

\theoremstyle{remark}

\usepackage[textsize=tiny]{todonotes}

\icmltitlerunning{Fully Decentralized Cooperative Multi-Agent Reinforcement Learning is A Context Modeling Problem}

\begin{document}

\twocolumn[
  \icmltitle{Fully Decentralized Cooperative Multi-Agent Reinforcement Learning \\ is A Context Modeling Problem}



  \icmlsetsymbol{equal}{*}

  \begin{icmlauthorlist}
    \icmlauthor{Chao Li}{njupt}
    \icmlauthor{Bingkun Bao}{njupt}
    \icmlauthor{Yang Gao}{nju}
  \end{icmlauthorlist}

  \icmlaffiliation{njupt}{School of Computer Science, Nanjing University of Posts and Telecommunications}
  \icmlaffiliation{nju}{School of Intelligent Science and Technology, Nanjing University}

  \icmlcorrespondingauthor{Bingkun Bao}{bingkunbao@njupt.edu.cn}
  \icmlcorrespondingauthor{Yang Gao}{gaoy@nju.edu.cn}

  \icmlkeywords{Machine Learning, ICML}

  \vskip 0.3in
]



\printAffiliationsAndNotice{}  

\begin{abstract}
  This paper studies fully decentralized cooperative multi-agent reinforcement learning, wherein each agent solely observes the states, its local actions, and the shared team rewards. For each agent, the lack of access to other agents' actions typically induces non-stationarity during value function updates and relative overgeneralization during value function estimation, which together impede effective cooperative policy learning. However, existing works fail to address both issues simultaneously, due to their inability to model the joint policy of other agents in a fully decentralized setting. To overcome this limitation, we propose a novel method termed Dynamics-Aware Context (DAC), which formalizes the task, as locally perceived by each agent, as a Contextual Markov Decision Process, and addresses both non-stationarity and relative overgeneralization through dynamics-aware context modeling. Specifically, DAC attributes the non-stationary local task dynamics to switches among unobserved contexts, each corresponding to a distinct joint policy of the other agents. Then, DAC models the step-wise dynamics distribution using latent variables, and refers to them as contexts. Accordingly, DAC learns a context-based value function to address the non-stationarity issue during per-agent value function updates. For value function estimation, an optimistic marginal value is derived to promote the selection of cooperative actions, thus addressing the relative overgeneralization issue. Empirically, we evaluate DAC across various cooperative tasks, and the results demonstrate that DAC consistently outperforms multiple baselines, highlighting its effectiveness.
\end{abstract}

\section{Introduction}
Multi-agent reinforcement learning (MARL) has emerged as a powerful technique for addressing cooperative tasks, driving substantial advancements in both algorithms (\emph{e.g.,} value decomposition methods~\cite{sunehag2017value,rashid2020monotonic,son2019qtran,wang2020qplex} and multi-agent policy gradient methods~\cite{lowe2017multi,foerster2018counterfactual,yu2022surprising,zhong2024heterogeneous}) and applications (\emph{e.g.,} traffic signal control~\cite{wang2020large}, autonomous vehicles~\cite{zhou2021smarts}, and vaccine allocation~\cite{hao2023gat}). Most of these advances depend on the centralized training (often with decentralized execution) paradigm, where global information, particularly the joint actions of all agents, is accessible during training. However, such direct access to other agents' actions is often unattainable in real-world domains. For example, in industrial automation scenarios, robots from different companies can not share action information due to privacy concerns or limited communication capabilities. In such cases, the \emph{fully decentralized learning} is required, where each agent learns based on only its own experiences, without access to other agents' actions, during both training and execution periods.

However, developing effective cooperative policies under the fully decentralized learning paradigm is challenging due to two critical challenges arising from the lack of access to other agents' actions. The first one is non-stationarity during per-agent value function updates. Since treating other agents as part of the environment, the local task dynamics perceived by each agent becomes non-stationary due to the evolving policies of other agents, undermining the convergence of value function updates~\cite{jiang2024fully}. The second one is relative overgeneralization, wherein the value estimations of each agent's local cooperative actions may be biased by other agents' exploratory or sub-optimal actions, hindering agents from selecting the optimal joint actions~\cite{matignon2012independent}. As a result, fully decentralized learning typically suffers from low efficiency and sub-optimal solutions, which limits the effectiveness of multi-agent cooperation.

Accordingly, we classify existing value-based MARL methods into two categories. The first category aims to address the non-stationarity issue. These methods either ensure stationary transition data by directly accessing other agents' actions~\cite{sunehag2017value,rashid2020monotonic}, designing multi-agent importance sampling weights, fingerprints~\cite{foerster2017stabilising}, and constructing ideal transition probabilities~\cite{jiang2022i2q}, or enable stationary policy updates based on alternative policy updates~\cite{su2024multi}. The second category primarily targets the relative overgeneralization issue, typically by rectifying the learned factored global action value function~\cite{son2019qtran,wang2020qplex,rashid2020weighted} or employing optimistic or lenient value function updates~\cite{lauer2000algorithm,matignon2007hysteretic,omidshafiei2017deep,panait2006lenient,wei2016lenient} to encourage the selection of optimal joint actions. Although both issues stem from the lack of information about other agents' action, existing fully decentralized MARL methods address either the non-stationarity issue or the relative overgeneralization problem in isolation, due to their inability to model the joint policy of other agents in a fully decentralized setting. Consequently, they fail to simultaneously resolve both challenges.

To overcome this limitation, we propose Dynamics-Aware Context (DAC), a novel method that formalizes the task, as locally perceived by each agent, as a Contextual Markov Decision Process (CMDP)~\cite{hallak2015contextual}, and further addresses both non-stationarity and relative overgeneralization from a dynamics-aware context modeling perspective. Specifically, DAC attributes the non-stationary local task dynamics of each agent to switches among unobserved contexts, each corresponding to a distinct joint policy of other agents. Then, drawing upon ideas from concept drift literature~\cite{lu2018learning}, DAC employs a sliding window alongside per-agent local trajectory to model step-wise dynamics distribution using latent variables. Since each agent's task dynamics is determined by the joint policy of other agents, these variables implicitly represent the other agents' joint policy at each time step. Accordingly, we refer to them as contexts and learn a context-based value function for each agent to address the non-stationarity issue in per-agent value function updates. During value function estimation, an optimistic marginal value is derived to discard the effects caused by other agents' sub-optimal actions, thereby facilitating the selection of optimal joint actions and addressing the relative overgeneralization problem. The above enables effective cooperative policy learning in a fully decentralized manner.

Empirically, we evaluate DAC across various cooperative tasks, including the matrix game, predator and prey, and the StarCraft Multi-Agent Challenge (SMAC)~\cite{samvelyan2019starcraft}. The results demonstrate significant performance 
gain against multiple baselines, validating DAC's effectiveness.

\section{Related Work}
In this section, we classify current value-based MARL methods into two categories and give a brief introduction to them.

The first category of works addresses the non-stationarity problem by constructing stationary transition data or policy updates. Specifically, canonical value decomposition methods such as VDN~\cite{sunehag2017value} and QMIX~\cite{rashid2020monotonic}) assume direct access to agents' joint actions to ensure stationary transitions during training. However, these methods often suffer from the relative overgeneralization issue due to the representational limitation of their learned factored global action value functions~\cite{gupta2021uneven}. For independent Q-learning (IQL)~\cite{tan1993multi} agents, the multi-agent importance sampling~\cite{foerster2017stabilising} technique assumes direct access to other agents' policies and calculates an importance weight to decay obsolete data during experience replay. Multi-agent fingerprints method~\cite{foerster2017stabilising} uses the training iteration numbers and exploration rates of other agents to estimate their policies, and augments per-agent local transitions with these estimates. However, such direct access to other agents' information assumed in above methods is unattainable in practice. I2Q~\cite{jiang2022i2q} addresses this by shaping ideal transition probabilities for each IQL agent in a fully decentralized manner, and guarantees convergence to the optimal joint policy. In comparison to I2Q's approach of addressing non-stationarity and relative overgeneralization by shaping ideal transition probabilities, this work aims for a novel context-aware framework to tackle both issues. In addition, to ensure stationary policy updates, MA2QL~\cite{su2024multi} enforces sequential policy updates among IQL agents. When an agent updates its policy, all others' policies remain fixed. Despite its promise, the sequential policy update typically leads to sample inefficiency, as it lacks the capacity for parallel policy updates.

The second category of works addresses the relative overgeneralization issue by rectifying the learned factored global action value function or updating per-agent local value function in optimistic or lenient manners. Specifically, for value decomposition methods with representational limitations, weighted QMIX~\cite{rashid2020weighted} places more weights on potentially optimal joint actions during value updates to exclusively recover correct value estimations for these critical actions. QTRAN~\cite{son2019qtran} and QPLEX~\cite{wang2020qplex} incorporate additional complementary terms to correct the discrepancy between the learned factored global action value functions and the true joint ones. For IQL, distributed Q-learning~\cite{lauer2000algorithm} employs an optimistic value function for each agent to discard the effect caused by other agents' exploratory or sub-optimal actions. This enables agents to identify and select their local cooperative actions, thus addressing the relative overgeneralization problem. However, due to the high optimism, distributed Q-learning is vulnerable to stochasticity. To avoid this issue, Hysteretic Q-learning
~\cite{matignon2007hysteretic,omidshafiei2017deep} updates per-agent value function using two learning rates for positive and negative temporal difference errors, respectively. Lenient learning~\cite{panait2006lenient,wei2016lenient} shifts from optimistic to standard value function update using gradually decreasing lenience. However, the optimistic value function update may cause value overestimation, particularly when the value function is approximated using neural networks. Moreover, the neglect of non-stationarity further hinders efficient policy learning.

In summary, existing fully decentralized MARL methods fail to address both non-stationarity and relative overgeneralization in a unified manner. To address this limitation, this work proposes to formalize the task perceived by each agent as a CMDP, and tackles both issues from a dynamics-aware context modeling perspective, thereby effectively promoting fully decentralized cooperative policy learning.

\section{Preliminary}
In this section, we formalize the task addressed by this work, and review the non-stationarity and relative overgeneralization issues in decentralized learning, as well as the CMDP.

\subsection{Multi-Agent Markov Decision Process}
\label{sec:mmdp}
We consider a cooperative multi-agent task that can be modeled as a Multi-Agent Markov Decision Process (MMDP) $\langle N,S,\boldsymbol{A},P,R,\gamma \rangle$, where $N=\{1,2,\ldots,n\}$ represents the agent set and $S$ is the state space. $\boldsymbol{A}=A^{1}\times A^{2}\times \ldots A^{n}$ is all agents' joint action space and $A^{i}$ denotes the local action space of agent $i\in N$. At each time step $t$, each agent $i$ observes the state $s_{t}\in S$ and selects its local action $a_{t}^{i}\in A^{i}$ according to its decentralized policy $\pi^{i}(a_{t}^{i}|s_{t})$. Given the joint action $\boldsymbol{a}_{t}=(a_{t}^{1}, a_{t}^{2}, \ldots, a_{t}^{n})$, the environment transits to the next state $s_{t+1}$ according to the state transition function $P(s_{t+1}|s_{t}, \boldsymbol{a}_{t})$, and provides a shared team reward $r_{t}$ based on the reward function $R(s_{t}, \boldsymbol{a}_{t})$. The goal is to learn the optimal joint policy $\boldsymbol{\pi}^{*}=(\pi^{1,*}, \pi^{2,*}, \ldots, \pi^{n,*})$, which maximizes the expected discounted cumulative return $\mathbb{E}_{\boldsymbol{\pi}, P}[\sum\nolimits_{t=0}^{\infty}\gamma^{t}r_{t}]$, where $\gamma$ denotes a discount factor.

We consider the fully decentralized learning, wherein each agent $i$ observes only the state $s_{t}$, its local action $a_{t}^{i}$, and the shared reward $r_{t}$. For each decentralized agent $i$, the perceived task can be modeled as a Markov Decision Process (MDP) $\langle S,A^{i},P^{i},R^{i},\gamma \rangle$ with dynamics defined as follows:
\begin{equation}
	\label{eq:local_mdp}
	\begin{aligned}
		P^{i}(s_{t+1}|s_{t},a_{t}^{i})=\sum\nolimits_{a_{t}^{-i}}\pi^{-i}(a_{t}^{-i}|s_{t})P(s_{t+1}|s_{t},\boldsymbol{a}_{t}), \\
		R^{i}(s_{t},a_{t}^{i})=\sum\nolimits_{a_{t}^{-i}}\pi^{-i}(a_{t}^{-i}|s_{t})R(s_{t}, \boldsymbol{a}_{t}),
	\end{aligned}
\end{equation}
where $\pi^{-i}$ and $a_{t}^{-i}$ respectively denote the joint policy and the joint action of other agents $-i$ except for agent $i$.

\textbf{Non-Stationarity.} As illustrated in Eq.~(\ref{eq:local_mdp}), each agent $i$'s local task dynamics, denoted by $P^{i}$ and $R^{i}$, depend on other agents $-i$'s joint policy $\pi^{-i}$. Since other agents continually change their policies, the local task dynamics of each agent $i$ becomes non-stationary. This non-stationarity undermines the convergence of per-agent value function updates.

\textbf{Relative Overgeneralization.} Due to the absence of other agents' action information, the value estimation of per-agent local cooperative actions may be biased by exploratory or sub-optimal actions taken by other agents. As a result, the sub-optimal joint actions are preferred over the optimal ones, a problem known as relative overgeneralization.

Specifically, for each decentralized agent $i$, its local value function $Q^{i}(s_{t},a_{t}^{i})$ can be regarded as a projection regarding the joint action value function $Q(s_{t}, a_{t}^{i}, a_{t}^{-i})$. IQL adheres to an average-based projection below:
\begin{equation}
	\label{eq:average_based_projection}
	\begin{aligned}
		Q^{i,\boldsymbol{\pi}}(s_{t},a_{t}^{i})=\sum\nolimits_{a_{t}^{-i}}\pi^{-i}(a_{t}^{-i}|s_{t})Q^{\boldsymbol{\pi}}(s_{t}, a_{t}^{i}, a_{t}^{-i}),
	\end{aligned}
\end{equation}
where $Q^{\boldsymbol{\pi}}(s_{t}, a_{t}^{i}, a_{t}^{-i})$ represents the joint action value function given a joint policy $\boldsymbol{\pi}=(\pi^{i}, \pi^{-i})$. As shown in Eq.~(\ref{eq:average_based_projection}), $Q^{i}$ following the average-based projection is easily affected by other agents' sub-optimal actions, thus suffering from the relative overgeneralization issue. In contrast, the maximum-based (optimistic) projection is defined below:
\begin{equation}
	\label{eq:optimistic_projection}
	\begin{aligned}
		Q^{i,\operatorname{opt}}(s_{t},a_{t}^{i})=\max\nolimits_{a_{t}^{-i}}Q^{*}(s_{t},a_{t}^{i},a_{t}^{-i}),
	\end{aligned}
\end{equation}
where $Q^{*}(s_{t},a_{t}^{i},a_{t}^{-i})$ is the joint action value function of an optimal joint policy $\boldsymbol{\pi}^{*}$. This optimistic projection assumes that other agents $-i$ always select their cooperative actions, thus eliminating their effects on agent $i$'s local value estimations. Both distributed Q-learning and hysteretic Q-learning approximate $Q^{i,\operatorname{opt}}(s_{t},a_{t}^{i})$ by an optimistic value function update. In contrast, our method estimates it by an optimistic marginal value derived from a context-based value function. We detail the distinction between them in Appendix.~\ref{sec:the_distinction_clarification}.

\subsection{Contextual Markov Decision Process}
\label{sec:cmdp}
A CMDP is often defined as a tuple $\langle \mathcal{C},S,A,M(c) \rangle$, where $\mathcal{C}$ denotes the context space, $S$ is the state space, and $A$ is the action space. For each context $c\in\mathcal{C}$, the function $M(c)$ specifies a MDP $\langle S,A,P^{c},R^{c},\gamma \rangle$. Thus, CMDP defines a family of MDPs that share the same state and action spaces but differ in the state transition and reward functions. In this work, we employ CMDP to model the non-stationary task dynamics locally perceived by each agent, where the contexts are associated with other agents' joint policies.

\section{Methodology}
In this section, we give a comprehensive introduction to our method, DAC. We begin by proposing the task formalization based on CMDP, and then delve into the process of modeling dynamics-aware contexts. Subsequently, for each agent, we learn a context-based value function and derive an optimistic marginal value, thereby addressing both non-stationarity and relative overgeneralization issues. Finally, we summarize the overall learning procedure of DAC.

\subsection{Task Formalization}
As detailed in Sec.~\ref{sec:mmdp}, the local task of each agent $i$ can be modeled as a MDP $\langle S,A^{i},P^{i},R^{i},\gamma \rangle$, where both the state transition function $P^{i}$ and the reward function $R^{i}$ depend on other agents $-i$' joint policy $\pi^{-i}$. Considering all possible $\pi^{-i}$, the task perceived by agent $i$ can be decomposed into a family of MDPs that share the same state and action spaces but differ in their transition and reward functions, with each MDP corresponding to a unique $\pi^{-i}$. By associating each context $c$ with a specific $\pi^{-i}$, we propose to formalize the perceived task of each agent $i$ as a CMDP, as defined below:
\begin{equation}
	\begin{aligned}
		\langle \mathcal{C},S,A^{i},M(c) \rangle, \ M(c): c\rightarrow\langle S,A^{i},P_{c}^{i},R_{c}^{i},\gamma \rangle,
	\end{aligned}
\end{equation}
where $S$ represents the state space and $A^{i}$ is the local action space of agent $i$. Note that the state transition function $P_{c}^{i}$ and the reward function $R_{c}^{i}$ are explicitly conditioned on the context $c\in\mathcal{C}$, each corresponding to a unique $\pi^{-i}$.

When each agent operates in a fully decentralized manner, under the above CMDP formalization, the task dynamics is determined by the underlying context, which corresponds to other agents' current joint policy. When an agent encounters different contexts at different time steps, the same states and local actions lead to different next states and rewards due to the distinct task dynamics. Consequently, the absence of context information hinders each agent from fully capturing the task dynamics, leading to the non-stationarity problem.

\emph{For each agent, this CMDP formalization attributes non-stationarity to switches between unobserved contexts, and provides a principled framework to address this problem by explicit context modeling.} The context can be instantiated as: (1) an estimate of other agents' current joint policy, or (2) a representation of the current agent's task dynamics distribution. By augmenting per-agent local transitions with the inferred contexts, the resulting transitions become stationary and enable stationary fully decentralized policy learning.

\begin{figure}[t]
	\centering
	\includegraphics[width=0.9\linewidth]{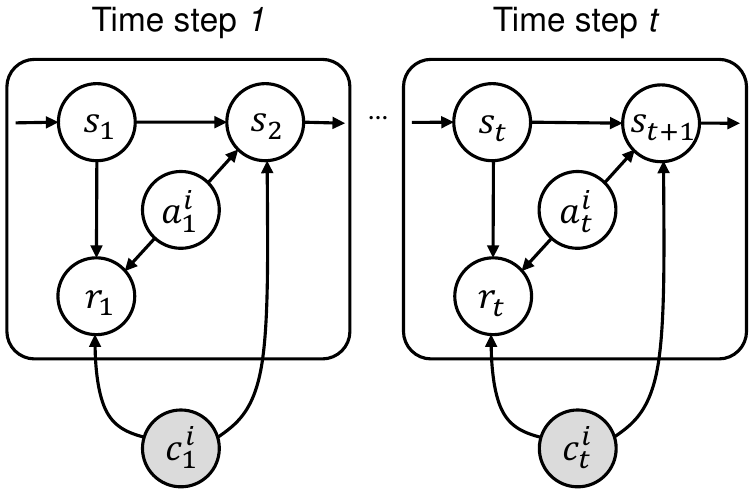}
	\caption{A general setting where the context changes every (or every few) time steps within agent $i$'s local task, depending on the update interval of other agents' joint policy. In the above plot, the local task dynamics is determined by $c_{1}^{i}$ at time step $1$ and by $c_{t}^{i}$ at time step $t$. Empty and solid circles represent observable and unobservable stochastic variables, respectively.}
	\label{fig:context_changes}
\end{figure}

We consider a general setting in which contexts change at every time step or over short time intervals, as depicted in Fig.~\ref{fig:context_changes}. In this setting, other agents update their joint policy at the same temporal scale, causing the context within the current agent's CMDP to shift correspondingly. In this work, we take a step toward explicitly modeling contexts within this setting, enabling principled handling of non-stationarity.


\subsection{Dynamics-Aware Context Modeling}
For each agent, we propose to represent its real-time local task dynamics distribution using latent variables, and refer to them as contexts. As other agents may update their joint policy every (or every few) time steps, the task dynamics distribution experienced along the current agent's local trajectory evolves on a comparable time scale. This parallels the setting of concept drift, in which the underlying data distribution evolves over time. To address such distributional shifts, maintaining a sliding window to hold the latest data within the data stream has proven effective in capturing the real-time data distribution, ensuring model adaptability and accuracy in dynamic settings~\cite{lu2018learning}.

Motivated by this insight, we cast the per-agent real-time task dynamics distribution modeling as a concept drift problem, and propose DAC as a solution. At first, DAC maintains a sliding window alongside per-agent local trajectory to hold the latest $k$ transitions. At time step $t$, the sliding window for agent $i$ is instantiated as the trajectory segment $\tau_{t-k+1:t}^{i}$, which contains transitions from time steps $t-k+1$ to $t$:
\begin{equation}
	\begin{aligned}
		\tau_{t-k+1:t}^{i}=&(s_{t-k+1},a_{t-k+1}^{i},r_{t-k+1},s_{t-k+2},\ldots, \\
		&s_{t},a_{t}^{i},r_{t},s_{t+1}).
	\end{aligned}
\end{equation}

For modeling the task dynamics distribution of $\tau_{t-k+1:t}^{i}$, we assume that this distribution can be represented by a latent variable $c_{t}^{i}$, and the underlying mapping from the trajectory segment to the variable adheres to an unknown probability distribution $p^{i}(c_{t}^{i}|\tau_{t-k+1:t}^{i})$. We learn an additional distribution $q^{i}(c_{t}^{i}|\tau_{t-k+1:t}^{i})$ to approximate it, and optimize this approximation by minimizing the KL-divergence between them (Detailed derivation can be found in Appendix.~\ref{sec:theoretical_derivation}):
\begin{equation}
	\label{eq:vae_inference}
	\begin{aligned}
		&D_{\operatorname{KL}}(q^{i}(c_{t}^{i}|\tau_{t-k+1:t}^{i})||p^{i}(c_{t}^{i}|\tau_{t-k+1:t}^{i}))\\
		&=\log p^{i}(\tau_{t-k+1:t}^{i})+D_{\operatorname{KL}}(q^{i}(c_{t}^{i}|\tau_{t-k+1:t}^{i})||p^{i}(c_{t}^{i})) \\
		&-\mathbb{E}_{q^{i}(c_{t}^{i}|\tau_{t-k+1:t}^{i})}\log p^{i}(\tau_{t-k+1:t}^{i}|c_{t}^{i}),
	\end{aligned}
\end{equation}
where $p^{i}(c_{t}^{i})$ denotes the true prior distribution of the latent variable, and $\log p^{i}(\tau_{t-k+1:t}^{i})$ represents the evidence that can be regarded as a constant. Based on Eq.~(\ref{eq:vae_inference}), to minimize the term $D_{\operatorname{KL}}(q^{i}(c_{t}^{i}|\tau_{t-k+1:t}^{i})||p^{i}(c_{t}^{i}|\tau_{t-k+1:t}^{i}))$, we aim to maximize the following equation:
\begin{equation}
	\label{eq:vae_optimize}
	\begin{aligned}
		\max &\underbrace{\mathbb{E}_{q^{i}(c_{t}^{i}|\tau_{t-k+1:t}^{i})}\log p^{i}(\tau_{t-k+1:t}^{i}|c_{t}^{i})}_{\textcircled{\small{1}}} \\
		&-\underbrace{D_{\operatorname{KL}}(q^{i}(c_{t}^{i}|\tau_{t-k+1:t}^{i})||p^{i}(c_{t}^{i}))}_{\textcircled{\small{2}}}.
	\end{aligned}
\end{equation}

In Eq.~(\ref{eq:vae_optimize}), term $\textcircled{\small{1}}$ represents the reconstruction likelihood that ensures the learned latent variable $c_{t}^{i}$ contains sufficient information of the trajectory segment $\tau_{t-k+1:t}^{i}$, and term ${\textcircled{\small{2}}}$ ensures the latent variable $c_{t}^{i}$ is close to its prior distribution. For optimizing term ${\textcircled{\small{1}}}$, we expand it as follows:
\begin{equation}
	\label{eq:term1_expand}
	\begin{aligned}
		p^{i}(&\tau_{t-k+1:t}^{i}|c_{t}^{i})=p(s_{t-k+1}) \\
		&\prod\nolimits_{h=t-k+1}^{t}p^{i}(a_{h}^{i}|s_{h})p^{i}(s_{h+1},r_{h}|s_{h},a_{h}^{i},c_{t}^{i}),
	\end{aligned}
\end{equation}
where the initial state distribution $p(s_{t-k+1})$ is determined by the environment and $p^{i}(a_{h}^{i}|s_{h})$ denotes agent $i$'s decentralized policy conditioned on the state. Therefore, we ignore these two components and rewrite Eq.~(\ref{eq:vae_optimize}) as follows:
\begin{equation}
	\label{eq:final_optimize}
	\begin{aligned}
		\max &\mathbb{E}_{q^{i}(c_{t}^{i}|\tau_{t-k+1:t}^{i})}\sum_{h=t-k+1}^{t}\log p^{i}(s_{h+1},r_{h}|s_{h},a_{h}^{i},c_{t}^{i})\\
		&-D_{\operatorname{KL}}(q^{i}(c_{t}^{i}|\tau_{t-k+1:t}^{i})||p^{i}(c_{t}^{i})).
	\end{aligned}
\end{equation}

Accordingly, DAC is capable of representing the real-time task dynamics distribution using the latent variable $c_{t}^{i}$, which is derived by the learned distribution $q^{i}(c_{t}^{i}|\tau_{t-k+1:t}^{i})$. These variables are then used as contexts to enable stationary policy learning, as detailed in subsequent sections.

\textbf{Context-based Value Function.} For each agent $i$, we learn a value function $Q^{i}(s_{t},a_{t}^{i},c_{t}^{i})$, which is additionally conditioned on the context $c_{t}^{i}$, besides the state $s_{t}$ and the local action $a_{t}^{i}$. The incorporation of contexts brings two benefits. On the one hand, the context is derived from each agent $i$'s local trajectory segment based on $q^{i}(c_{t}^{i}|\tau_{t-k+1:t}^{i})$, enabling fully decentralized policy learning via $Q^{i}(s_{t},a_{t}^{i},c_{t}^{i})$. On the other hand, the context alleviates the non-stationarity caused by other agents' evolving joint policy, therefore facilitating stationary update of $Q^{i}(s_{t},a_{t}^{i},c_{t}^{i})$. Based on the augmented transition $(s_{t},a_{t}^{i},r_{t},c_{t}^{i},s_{t+1},c_{t+1}^{i})$, $Q^{i}$ is updated as below:
\begin{equation}
	\label{eq:context_value_function}
	\begin{aligned}
		&\mathcal{L}_{\operatorname{C}}(\theta^{i})=\mathbb{E}_{(s_{t},a_{t}^{i},c_{t}^{i},r_{t},s_{t+1},c_{t+1}^{i})\sim\mathcal{D}^{i}} \\
		&(r_{t}+\gamma\max_{a_{t+1}^{i}}Q^{i}(s_{t+1},a_{t+1}^{i},c_{t+1}^{i})-  Q^{i}(s_{t},a_{t}^{i},c_{t}^{i}))^{2},
	\end{aligned}
\end{equation}
where $Q^{i}$ is parameterized by $\theta^{i}$ and we sample transitions from agent $i$'s own replay buffer $\mathcal{D}^{i}$ to conduct the update.

\textbf{Optimistic Marginal Value.} There are still two problems during value function estimation: (1) The relative overgeneralization hinders agents from identifying and selecting their local cooperative actions; and (2) Each agent can not select actions based on the context-based value function, where the context based on real-time task dynamics modeling is available only when the current transition (\emph{i.e.,} $(s_{t},a_{t}^{i},r_{t},s_{t+1})$) is finished. To deal with these two problems, we propose the optimistic marginal value for each agent, as defined below:
\begin{equation}
	\label{eq:optimistic_marginal_value}
	\begin{aligned}
		\phi^{i}(s_{t},a_{t}^{i})=\max_{c\in\mathcal{C}}Q^{i}(s_{t},a_{t}^{i},c),
	\end{aligned}
\end{equation}
where we shape marginal value functions of per-agent local actions using the maximum context-based value estimations across all possible contexts $c\in \mathcal{C}$. Eq.~(\ref{eq:optimistic_marginal_value}) adheres to an optimistic belief that other agents always select their cooperative actions, such that the marginal value of per-agent local action can attain the maximum value $\max_{c\in\mathcal{C}}Q^{i}(s_{t},a_{t}^{i},c)$. Under this formulation, the optimistic marginal value can discard the effects caused by other agents' exploratory or sub-optimal actions, thus enabling accurate identifications and selections of per-agent local cooperative actions. As a result, this marginal value addresses relative overgeneralization and promotes multi-agent cooperation. Moreover, for each agent $i$, the resulting value function $\phi^{i}(s_{t}, a_{t}^{i})$ induces a decentralized policy $\pi^{i}(a_{t}^{i}|s_{t})$ that depends solely on the state $s_{t}$, satisfying the assumption in Eq.~(\ref{eq:term1_expand}) and Eq.~(\ref{eq:final_optimize}).

\subsection{Overall Learning Procedure}
\textbf{Discrete Context.} In Eq.~(\ref{eq:optimistic_marginal_value}), to efficiently enumerate all possible contexts, we propose to construct a discrete context space using a VAE-like network~\cite{kingma2013auto}. As depicted in Fig.~\ref{fig:dac_architecture_appendix}, we implement the mapping function $q^{i}(c_{t}^{i}|\tau_{t-k+1:t}^{i})$ as the encoder, which takes as input a sliding window containing the most recent $k$ transitions and outputs the logits of a categorical context distribution. The decoder then reconstruct the corresponding task dynamics.

To shape discrete contexts, we consider two aspects. First, the learned context distribution is encouraged to match a discrete prior $p^{i}(c_{t}^{i})$, defined as a uniform categorical distribution. Thus, term ${\textcircled{\small{2}}}$ in Eq.~(\ref{eq:vae_optimize}) is achieved by the loss:
\begin{equation}
	\label{eq:kl_loss_function}
	\begin{aligned}
		\mathcal{L}_{\operatorname{KL}}&(\omega_{\operatorname{e}}^{i})=D_{\operatorname{KL}}(q^{i}(c_{t}^{i}|\tau_{t-k+1:t}^{i})||p^{i}(c_{t}^{i})) \\
		&=\sum_{j=1}^{m}q_{j}^{i}(c_{t}^{i}|\tau_{t-k+1:t}^{i})\log\frac{q_{j}^{i}(c_{t}^{i}|\tau_{t-k+1:t}^{i})}{1/m},
	\end{aligned}
\end{equation}
where the encoder is parameterized by $\omega_{\operatorname{enc}}^{i}$, $m$ denotes the number of discrete context categories (\emph{i.e.,} the dimension of context), and $q_{j}^{i}$ represents the probability that $c_{t}^{i}$ belongs to the $j$-th category. This probability is obtained by applying a softmax function to the encoder outputs. 

Second, to ensure discrete context sampling while retaining differentiable, we employ the Gumbel-Softmax~\cite{jang2016categorical} to the encoder outputs. Given a sampled context $c_{t}^{i}$, the decoder reconstructs the local task dynamics by maximizing $\sum_{h=t-k+1}^{t}\log p^{i}(s_{h+1},r_{h}|s_{h},a_{h}^{i},c_{t}^{i})$, which corresponds to the following reconstruction loss:
\begin{equation}
	\label{eq:vae_loss_function}
	\begin{aligned}
		&\mathcal{L}_{\operatorname{REC}}(\omega_{\operatorname{e}}^{i},\omega_{\operatorname{d}}^{i})=\mathbb{E}_{\tau_{t-k+1:t}^{i}\sim\mathcal{D}^{i}, c_{t}^{i}\sim q^{i}(c_{t}^{i}|\tau_{t-k+1:t}^{i})}\sum_{h=t-k+1}^{t} \\
		&[(s_{h+1}-f_{\operatorname{s}}(s_{h},a_{h}^{i},c_{t}^{i}))^{2} + (r_{h}-f_{\operatorname{r}}(s_{h},a_{h}^{i},c_{t}^{i}))^{2}],
	\end{aligned}
\end{equation}
where the decoder consists of two separate networks, $f_{\operatorname{s}}$ and $f_{\operatorname{r}}$, used to respectively predict the next states and rewards. $\omega_{\operatorname{d}}^{i}$ denotes the decoder parameters. The Gumbel-Softmax enables differentiable training of the encoder. As a result, the encoder and decoder are jointly optimized using the loss function $\mathcal{L}_{\operatorname{V}}=\mathcal{L}_{\operatorname{REC}}+\beta\mathcal{L}_{\operatorname{KL}}$, where $\beta$ is a scaling factor.

\textbf{Augmented Exploration.} Maintaining accurate estimations of the context-based value function requires comprehensive coverage of per-agent local task dynamics during training. To this end, we introduce an augmented exploration scheme that facilitates the collection of diverse transitions. This is implemented by applying a hierarchical $\epsilon$-greedy strategy to both context and action sampling. Specifically, we define the marginal value $q^{i}(s_{t},a_{t}^{i})$ for each agent $i$ as follows:
\begin{equation}
	\label{eq:context_sampling}
	\begin{aligned}
		q^{i}(s_{t},a_{t}^{i})=\begin{cases} \phi^{i}(s_{t}, a_{t}^{i}) & p=1-\epsilon \\ Q^{i}(s_{t},a_{t}^{i}, c), c\in \mathcal{U}(\mathcal{C}) & p=\epsilon \end{cases},
	\end{aligned}
\end{equation}
where $\mathcal{U}(\mathcal{C})$ denotes a uniform distribution over the context space $\mathcal{C}$ and $p$ represents the probability. In this formulation, the context $c$ is sampled via an $\epsilon$-greedy policy to compute the marginal value $q^{i}(s_{t},a_{t}^{i})$. Then, agent $i$ samples its local action using an $\epsilon$-greedy policy with respect to $q^{i}(s_{t},a_{t}^{i})$.

The context-based value function $Q^{i}(s_{t}, a_{t}^{i}, c)$ essentially maintains a distribution of value estimations over per-agent local state-action pairs. Eq.~(\ref{eq:context_sampling}) naturally incorporates this distribution into the exploration process. This augmented exploration scheme is inspired by Bootstrapped DQN~\cite{osband2016deep}, which demonstrates that action selection derived from a set of distinct value functions induces more efficient exploration. We empirically demonstrate in Sec.~\ref{sec:exp} that this scheme significantly enhances learning efficiency. More algorithmic details can be found in Appendix.~\ref{sec:algorithmic_details}.

\begin{figure*}[t]
	\centering
	\begin{subfigure}[b]{0.48\textwidth}
		\centering
		\includegraphics[width=0.9\textwidth]{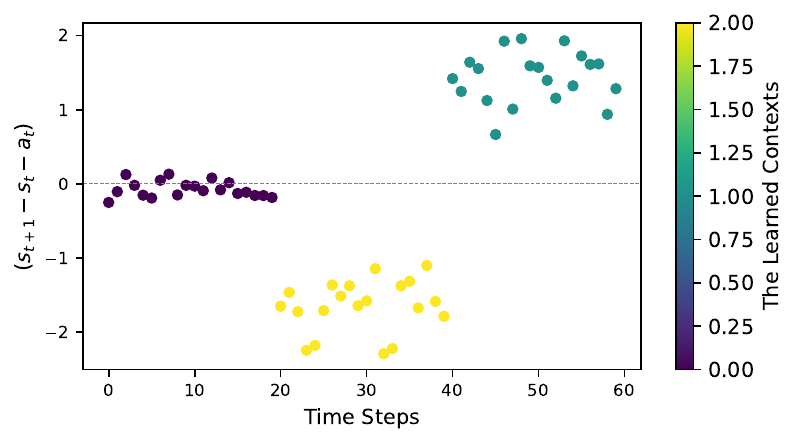}
		\label{fig:didactic_exp_1}
	\end{subfigure}
	\
	\begin{subfigure}[b]{0.48\textwidth}
		\centering
		\includegraphics[width=0.86\textwidth]{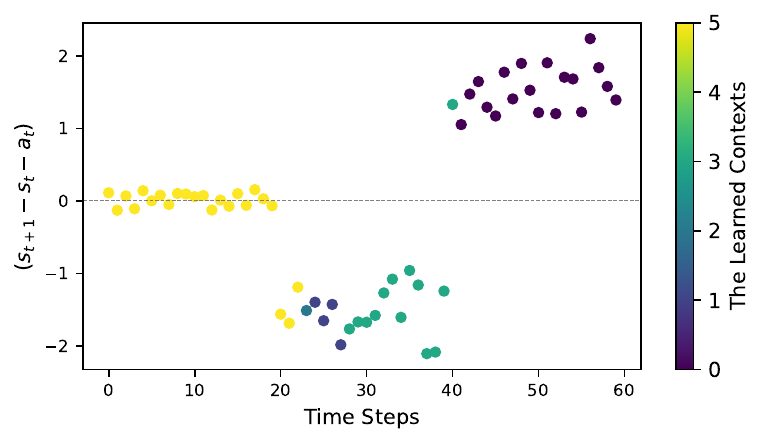}
		\label{fig:didactic_exp_2}
	\end{subfigure}
	\vspace{-5pt}
	\caption{Visualization of the learned contexts in Switching Wind. The left plot corresponds to the setting with $k=1$ and $m=3$, while the right plot corresponds to the setting with $k=5$ and $m=6$.}
	\label{fig:didactic_exp}
\end{figure*}

\section{Experiment}
\label{sec:exp}
In this section, we design experiments to answer questions below. (1) Can DAC capture non-stationary task dynamics using the learned discrete contexts? (See Sec.~\ref{subsec:didactic_example}) (2) Can DAC benefit fully decentralized cooperative policy learning by addressing both non-stationarity and relative overgeneralization issues? (See Sec.~\ref{subsec:comparison_results}) (3) If so, which component contributes the most to its performance gain? (See Sec.~\ref{subsec:ablation_study})

\subsection{Didactic Example}
\label{subsec:didactic_example}
For question (1), we introduce a didactic single-agent task, named Switching Wind, where a single agent moves along a line and an external wind (modeled as Gaussian noise with a periodically shifting mean) changes the transition dynamics. This abstracts multi-agent settings, where changes in other agents' policies induce non-stationarity in each agent's local dynamics. Formally, the state $s_{t}\in \mathbb{R}$ is the agent's position and the action $a_{t}\in[-1, 1]$ is a movement scalar. The state transition follows $s_{t+1}=s_{t}+a_{t}+\epsilon_{t}$, where $\epsilon_{t}$ depends on the task mode: Calm ($\epsilon\sim \mathcal{N}(0, 0.1)$), Headwind ($\epsilon\sim \mathcal{N}(-1.5, 0.3)$), or Tailwind ($\epsilon\sim \mathcal{N}(1.5, 0.3)$). An episode lasts 60 steps and the task mode switches every 20 steps.

We train the VAE-like network using an episode of randomly sampled transitions, and visualize the learned discrete contexts. As shown in Fig.~\ref{fig:didactic_exp}, with a single-transition window ($k=1$) and 3 contexts ($m=3$), the learned contexts align uniquely with the underlying task modes, demonstrating the model's ability to capture non-stationary dynamics. When longer sliding windows ($k>1$) are used, the empirical transition distribution varies smoothly across mode boundaries, requiring a larger number of discrete contexts. Accordingly, with $k=5$ and $m=6$, transitions near mode-switching points are assigned complementary context labels, reflecting mixed dynamics within the sliding window.

Compared with single-transition estimation ($k=1$), sliding-window-based estimation incurs a delay in capturing non-stationarity but improves robustness, wherein aggregating multiple transitions reduces sensitivity to stochastic noise and yields more stable context shifts. In contrast, the single-transition estimation can react immediately but is prone to outliers, leading to oscillatory and inefficient context assignments. We empirically validate the effectiveness of sliding-window-based estimation in subsequent comparisons.

\begin{table*}[t]
	\centering
	\begin{tabular}{ccc}
		\begin{subtable}[t]{0.23\linewidth}
			\centering
			\small
			\scalebox{1.0}{
				\begin{tabular}{|c|c|c|c|}
					\hline 
					\diagbox[dir=SE]{$A^1$}{$A^2$} &  $a^{1}$ & $a^{2}$ & $a^{3}$ \\ 
					\hline 
					$a^{1}$ &  8 &  -12 & -12  \\ 
					\hline
					$a^{2}$ &  -12 & 6 & 0 \\ 
					\hline
					$a^{3}$ &  -12 & 0 & 6 \\ 
					\hline 
				\end{tabular}
			}
			\caption{Payoff matrix.}
			\label{table:matrix_game}
		\end{subtable}
		&
		\begin{subtable}[t]{0.33\linewidth}
			\centering
			\small
			\scalebox{1.0}{
				\begin{tabular}{|c|c|c|c|c|}
					\hline 
					\diagbox[dir=SE]{$A^1$}{$c$} &  $0$ & $1$ & $2$ & $3$ \\ 
					\hline 
					$a^{1}(8.03)$ &  $\bcancel{-0.7}$ &  $-11$ & $8.03$ & $\bcancel{-5.0}$  \\ 
					\hline
					$a^{2}(5.96)$ &  $0.01$ & $-12$ & $5.96$ & $0.00$ \\ 
					\hline
					$a^{3}(5.95)$ &  $0.06$ & $-12$ & $5.95$ & $-0.0$ \\ 
					\hline 
				\end{tabular}
			}
			\caption{$Q^{1}(s,a,c)$ and $\phi^{1}(s,a)$.}
			\label{table:matrix_game_dac_agent1}
		\end{subtable}
		&
		\begin{subtable}[t]{0.33\linewidth}
			\centering
			\small
			\scalebox{1.0}{
				\begin{tabular}{|c|c|c|c|c|}
					\hline 
					\diagbox[dir=SE]{$A^2$}{$c$} &  $0$ & $1$ & $2$ & $3$ \\ 
					\hline 
					$a^{1}(8.03)$ &  $\bcancel{-0.7}$ & $-11$ & $8.03$ & $\bcancel{-5.0}$  \\ 
					\hline
					$a^{2}(5.96)$ &  $0.01$ & $-12$ & $5.96$ & $0.01$ \\ 
					\hline
					$a^{3}(5.94)$ &  $0.07$ & $-12$ & $5.94$ & $-0.0$ \\ 
					\hline 
				\end{tabular}
			}
			\caption{$Q^{2}(s,a,c)$ and $\phi^{2}(s,a)$.}
			\label{table:matrix_game_dac_agent2}
		\end{subtable}
	\end{tabular}
	\caption{The payoff matrix and value functions learned by DAC. We set the number of discrete contexts to $4$. The context-based value functions $Q^{1}(s,a,c)$ and $Q^{2}(s,a,c)$ for all contexts $c$ are presented in (b) and (c), respectively. The optimistic marginal values, $\phi^{1}(s,a)$ and $\phi^{2}(s,a)$, appear in the first column. $Q^{i}(s,a,c)$ with unattainable value is marked by $\bcancel{Q^{i}(s,a,c)}$.}
\end{table*}

\begin{figure*}[t]
	\centering
	\includegraphics[width=0.99\linewidth]{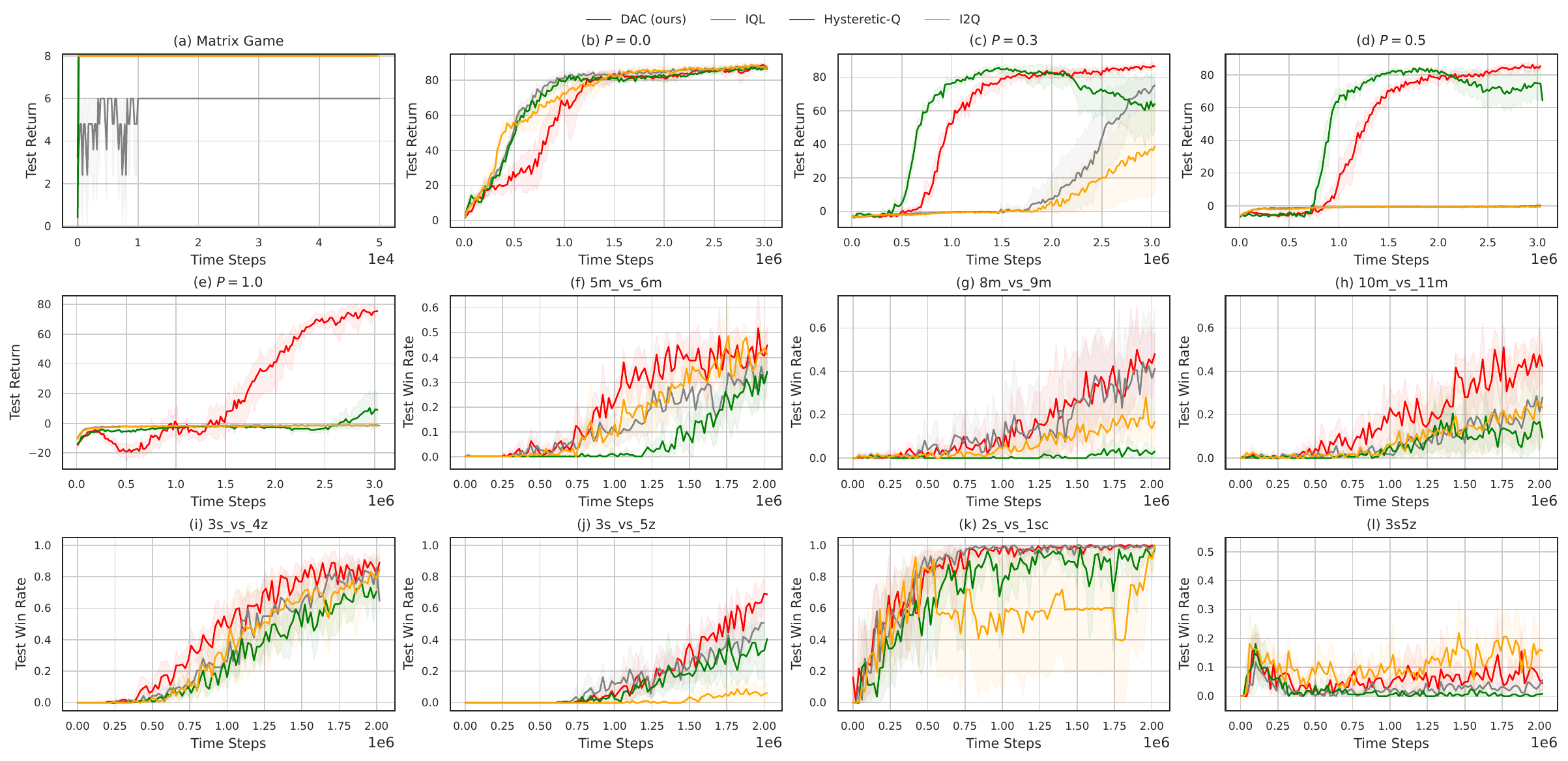}
	\caption{Comparison results in the matrix game, predator and prey, and several SMAC maps.}
	\label{fig:comparison_results}
\end{figure*}

\subsection{Comparison Results}
\label{subsec:comparison_results}
For addressing question (2), we compare our method against multiple fully decentralized value-based MARL baselines, including IQL~\cite{tan1993multi}, Hysteretic Q-learning~\cite{matignon2007hysteretic}, and I2Q~\cite{jiang2022i2q}, on the matrix game, predator and prey, and the SMAC benchmarks. 

\textbf{Matrix Game.} We begin by evaluating all methods on a matrix game. As shown in Tab.~\ref{table:matrix_game}, two agents within this game need to select the optimal joint action $(a^{1},a^{1})$ to achieve the best reward $+8$. However, under the fully decentralized learning paradigm, each agent maintains higher value estimations regarding its local actions $a^{2}$ and $a^{3}$, when the other agent selects actions uniformly at random. This gives rise to the relative overgeneralization problem where sub-optimal joint actions are preferred over the optimal one.

Fig.~\ref{fig:comparison_results} (a) shows the comparison results of all methods in the matrix game. One can observe that IQL struggles in the sub-optimal joint actions with $+6$ rewards, demonstrating that the average-based projection (Eq.~(\ref{eq:average_based_projection})) followed by IQL is susceptible to the relative overgeneralization. In contrast, DAC provides each agent with an optimistic marginal value, and accordingly discards the effects caused by other agents' sub-optimal actions. As a result, DAC succeeds in selecting the optimal joint action with $+8$ reward. This also applies to Hysteretic Q-learning, which adheres to an optimistic value function update and demonstrates efficiency in simple tasks. However, when faced with complex cooperative tasks, such optimistic value function update often leads to overestimation for value function approximated by neural networks, which leads to poor performance. We validate this insight in subsequent experiments. Similarly, I2Q shapes ideal transitions by implicitly assuming other agents follow cooperative policies, which adheres to an optimistic belief, and learning policies on these transitions leads to the optimal joint policy. 

Furthermore, to analyze the representational capabilities of the contexts, we present the per-agent context-based value function and optimistic marginal value learned by DAC. As depicted in Tab.~\ref{table:matrix_game_dac_agent1} and Tab.~\ref{table:matrix_game_dac_agent2}, for agent $1$ or $2$, the context-based value function accurately approximates the rewards of all possible joint actions, which validates the effectiveness of dynamics-aware contexts in representing other agents' joint policies. Moreover, the optimistic marginal values of per-agent local actions satisfy the optimistic property, \emph{i.e.,} which equal to the highest rewards only achieved when other agents select their cooperative actions. Accordingly, each agent can accurately select its local cooperative actions and the efficient selection of optimal joint actions is achieved.

\textbf{Predator and Prey.} To further assess the effectiveness of DAC in addressing relative overgeneralization, we adopt the modified predator-prey~\cite{son2019qtran}, where two predators receive a team reward of $+1$ when they simultaneously capture the single prey, otherwise $-P$ penalty for sole hunting. $P$ controls the degree of relative overgeneralization. 

We consider four settings $P=0.0, 0.3, 0.5, 1.0$. In general, a small penalty ($P=0.0$) does not induce relative overgeneralization, and thereby all methods succeed in learning cooperative prey-capturing policies with high rewards, as illustrated in Fig.~\ref{fig:comparison_results} (b). As $P$ increases and relative overgeneralization becomes more pronounced, IQL fails to learn effective policies for $P=0.3$, $0.5$, and $1.0$. In contrast, Hysteretic Q-learning and DAC achieve high rewards under moderate penalties ($P=0.3$ and $0.5$), benefiting from their ability to select cooperative actions based on optimistic value estimations. However, Hysteretic Q-learning exhibits degraded asymptotic performance due to the overestimation issue. In the most challenging scenario with $P=1.0$, only DAC succeeds in learning an effective policy yielding high rewards, while all other baselines fail. We attribute DAC's consistent superiority to its decoupled learning framework, which separately models the stationary context-based value function and the optimistic marginal value. By contrast, I2Q performs poorly for $P=0.3$, $0.5$, and $1.0$, suggesting that shaping ideal transition probabilities is difficult in practice.

\textbf{SMAC.} To evaluate the scalability of DAC to more complex tasks, we test all methods on a set of SMAC maps, including 5m\_vs\_6m, 8m\_vs\_9m, 10m\_vs\_11m, 3s\_vs\_4z, 3s\_vs\_5z, 2s\_vs\_1sc, and 3s5z. As depicted in Fig.~\ref{fig:comparison_results} (f) - (l), three key observations emerge. (1) On moderately difficult maps (\emph{i.e.,} 5m\_vs\_6m, 8m\_vs\_9m, 10m\_vs\_11m, 3s\_vs\_4z, 3s\_vs\_5z), DAC consistently outperforms the baselines. This advantage stems from its efficient selection of cooperative joint actions based on the learned optimistic marginal values. In contrast, Hysteretic Q-learning performs poorly, as overestimation and non-stationarity are exacerbated in complex tasks. IQL similarly suffers from degraded performance on most maps due to non-stationarity issue. (2) On simpler maps such as 2s\_vs\_1sc, DAC achieves performance comparable to IQL, where dense reward signals facilitate frequent cooperative behaviors and thus enable efficient policy updates for most methods. Consequently, the benefit brought by DAC is not obvious. (3) On particularly challenging maps in which all methods struggle, such as 3s5z, DAC also fails to converge to effective policies. We hypothesis that this limitation arises from inaccurate estimation of per-agent context-based value function, suggesting that more effective exploration methods are necessitated to adequately cover each agent's local task dynamics. In addition, I2Q performs poorly across all maps, which we attribute to its inability to efficiently shape ideal transition probabilities under partial observability.

\subsection{Ablation Study}
\label{subsec:ablation_study}
To examine the effects of DAC's components: (1) the sliding window length ($k$), (2) the number of discrete contexts ($m$), and (3) the KL loss ($\beta$), we set $k$ to 5, 10, 15, 20, $m$ to 5, 10, 15, 20, and $\beta$ to 0.001, 0.01, 0.1, 0.0 as multiple baselines. To evaluate (4) the augmented exploration, we introduce a variant, DAC\_$\phi^{i}$, in which agents select actions solely based on
$\phi^{i}(s_{t}, a_{t}^{i})$. Refer to Appendix.~\ref{sec:ablation_study_appendix} for detailed analyses.

\section{Conclusion}
This paper presents DAC as a unified framework to address both non-stationarity and relative overgeneralization issues for fully decentralized cooperative MARL. DAC formalizes the local task, as perceived by each agent, as a CMDP, and shapes contexts by modeling step-wise dynamics distribution. Then, DAC learns a context-based value function for each agent to enable stationary policy updates, and derives an optimistic marginal value to encourage the selection of cooperative joint actions. Extensive experiments on various cooperative tasks empirically validate its effectiveness.

\textbf{Limitation and Future Work.} We identify three limitations that warrant further investigation. First, DAC necessitates a comprehensive coverage of each agent's local task dynamics, yet the proposed hierarchical exploration over both context and action spaces becomes ineffective in highly challenging tasks. This can be mitigated by combining DAC with more efficient coordinated exploration techniques. Second, DAC uses a fixed-length sliding window to model non-stationary dynamics distribution, which often incurs a delay in capturing distribution shifts. We plan to explore adaptive sliding window and alternative dynamics modeling methods. Third, the computation of optimistic marginal values becomes increasingly costly as the context space grows. This issue can be alleviated by adopting sampling-based heuristic search methods to approximate the maxima and accordingly enable moderate complexity. We leave them as our future work.


\nocite{langley00}

\bibliography{reference.bib}

\begin{thebibliography}{30}
\providecommand{\natexlab}[1]{#1}
\providecommand{\url}[1]{\texttt{#1}}
\expandafter\ifx\csname urlstyle\endcsname\relax
  \providecommand{\doi}[1]{doi: #1}\else
  \providecommand{\doi}{doi: \begingroup \urlstyle{rm}\Url}\fi

\bibitem[Foerster et~al.(2017)Foerster, Nardelli, Farquhar, Afouras, Torr,
  Kohli, and Whiteson]{foerster2017stabilising}
Foerster, J., Nardelli, N., Farquhar, G., Afouras, T., Torr, P.~H., Kohli, P.,
  and Whiteson, S.
\newblock Stabilising experience replay for deep multi-agent reinforcement
  learning.
\newblock In \emph{International conference on machine learning}, pp.\
  1146--1155. PMLR, 2017.

\bibitem[Foerster et~al.(2018)Foerster, Farquhar, Afouras, Nardelli, and
  Whiteson]{foerster2018counterfactual}
Foerster, J., Farquhar, G., Afouras, T., Nardelli, N., and Whiteson, S.
\newblock Counterfactual multi-agent policy gradients.
\newblock In \emph{Proceedings of the AAAI conference on artificial
  intelligence}, volume~32, 2018.

\bibitem[Gupta et~al.(2021)Gupta, Mahajan, Peng, B{\"o}hmer, and
  Whiteson]{gupta2021uneven}
Gupta, T., Mahajan, A., Peng, B., B{\"o}hmer, W., and Whiteson, S.
\newblock Uneven: Universal value exploration for multi-agent reinforcement
  learning.
\newblock In \emph{International Conference on Machine Learning}, pp.\
  3930--3941. PMLR, 2021.

\bibitem[Hallak et~al.(2015)Hallak, Di~Castro, and
  Mannor]{hallak2015contextual}
Hallak, A., Di~Castro, D., and Mannor, S.
\newblock Contextual markov decision processes.
\newblock \emph{arXiv preprint arXiv:1502.02259}, 2015.

\bibitem[Hao et~al.(2023)Hao, Huang, Feng, Yuan, and Li]{hao2023gat}
Hao, Q., Huang, W., Feng, T., Yuan, J., and Li, Y.
\newblock Gat-mf: Graph attention mean field for very large scale multi-agent
  reinforcement learning.
\newblock In \emph{Proceedings of the 29th ACM SIGKDD Conference on Knowledge
  Discovery and Data Mining}, pp.\  685--697, 2023.

\bibitem[Jang et~al.(2016)Jang, Gu, and Poole]{jang2016categorical}
Jang, E., Gu, S., and Poole, B.
\newblock Categorical reparameterization with gumbel-softmax.
\newblock \emph{arXiv preprint arXiv:1611.01144}, 2016.

\bibitem[Jiang \& Lu(2022)Jiang and Lu]{jiang2022i2q}
Jiang, J. and Lu, Z.
\newblock I2q: A fully decentralized q-learning algorithm.
\newblock \emph{Advances in Neural Information Processing Systems},
  35:\penalty0 20469--20481, 2022.

\bibitem[Jiang et~al.(2024)Jiang, Su, and Lu]{jiang2024fully}
Jiang, J., Su, K., and Lu, Z.
\newblock Fully decentralized cooperative multi-agent reinforcement learning: A
  survey.
\newblock \emph{arXiv preprint arXiv:2401.04934}, 2024.

\bibitem[Kingma \& Welling(2014)Kingma and Welling]{kingma2013auto}
Kingma, D.~P. and Welling, M.
\newblock Auto-encoding variational bayes.
\newblock In \emph{International Conference on Learning Representations}, 2014.
\newblock URL \url{http://arxiv.org/abs/1312.6114}.

\bibitem[Lauer \& Riedmiller(2000)Lauer and Riedmiller]{lauer2000algorithm}
Lauer, M. and Riedmiller, M.~A.
\newblock An algorithm for distributed reinforcement learning in cooperative
  multi-agent systems.
\newblock In \emph{Proceedings of the seventeenth international conference on
  machine learning}, pp.\  535--542, 2000.

\bibitem[Lowe et~al.(2017)Lowe, Wu, Tamar, Harb, Pieter~Abbeel, and
  Mordatch]{lowe2017multi}
Lowe, R., Wu, Y., Tamar, A., Harb, J., Pieter~Abbeel, O., and Mordatch, I.
\newblock Multi-agent actor-critic for mixed cooperative-competitive
  environments.
\newblock \emph{Advances in neural information processing systems}, 30, 2017.

\bibitem[Lu et~al.(2018)Lu, Liu, Dong, Gu, Gama, and Zhang]{lu2018learning}
Lu, J., Liu, A., Dong, F., Gu, F., Gama, J., and Zhang, G.
\newblock Learning under concept drift: A review.
\newblock \emph{IEEE transactions on knowledge and data engineering},
  31\penalty0 (12):\penalty0 2346--2363, 2018.

\bibitem[Matignon et~al.(2007)Matignon, Laurent, and
  Le~Fort-Piat]{matignon2007hysteretic}
Matignon, L., Laurent, G.~J., and Le~Fort-Piat, N.
\newblock Hysteretic q-learning: an algorithm for decentralized reinforcement
  learning in cooperative multi-agent teams.
\newblock In \emph{2007 IEEE/RSJ International Conference on Intelligent Robots
  and Systems}, pp.\  64--69. IEEE, 2007.

\bibitem[Matignon et~al.(2012)Matignon, Laurent, and
  Le~Fort-Piat]{matignon2012independent}
Matignon, L., Laurent, G.~J., and Le~Fort-Piat, N.
\newblock Independent reinforcement learners in cooperative markov games: a
  survey regarding coordination problems.
\newblock \emph{The Knowledge Engineering Review}, 27\penalty0 (1):\penalty0
  1--31, 2012.

\bibitem[Omidshafiei et~al.(2017)Omidshafiei, Pazis, Amato, How, and
  Vian]{omidshafiei2017deep}
Omidshafiei, S., Pazis, J., Amato, C., How, J.~P., and Vian, J.
\newblock Deep decentralized multi-task multi-agent reinforcement learning
  under partial observability.
\newblock In \emph{International Conference on Machine Learning}, pp.\
  2681--2690. PMLR, 2017.

\bibitem[Osband et~al.(2016)Osband, Blundell, Pritzel, and
  Van~Roy]{osband2016deep}
Osband, I., Blundell, C., Pritzel, A., and Van~Roy, B.
\newblock Deep exploration via bootstrapped dqn.
\newblock \emph{Advances in neural information processing systems}, 29, 2016.

\bibitem[Panait et~al.(2006)Panait, Sullivan, and Luke]{panait2006lenient}
Panait, L., Sullivan, K., and Luke, S.
\newblock Lenient learners in cooperative multiagent systems.
\newblock In \emph{Proceedings of the fifth international joint conference on
  Autonomous agents and multiagent systems}, pp.\  801--803, 2006.

\bibitem[Rashid et~al.(2020{\natexlab{a}})Rashid, Farquhar, Peng, and
  Whiteson]{rashid2020weighted}
Rashid, T., Farquhar, G., Peng, B., and Whiteson, S.
\newblock Weighted qmix: Expanding monotonic value function factorisation for
  deep multi-agent reinforcement learning.
\newblock \emph{Advances in neural information processing systems},
  33:\penalty0 10199--10210, 2020{\natexlab{a}}.

\bibitem[Rashid et~al.(2020{\natexlab{b}})Rashid, Samvelyan, De~Witt, Farquhar,
  Foerster, and Whiteson]{rashid2020monotonic}
Rashid, T., Samvelyan, M., De~Witt, C.~S., Farquhar, G., Foerster, J., and
  Whiteson, S.
\newblock Monotonic value function factorisation for deep multi-agent
  reinforcement learning.
\newblock \emph{The Journal of Machine Learning Research}, 21\penalty0
  (1):\penalty0 7234--7284, 2020{\natexlab{b}}.

\bibitem[Samvelyan et~al.(2019)Samvelyan, Rashid, De~Witt, Farquhar, Nardelli,
  Rudner, Hung, Torr, Foerster, and Whiteson]{samvelyan2019starcraft}
Samvelyan, M., Rashid, T., De~Witt, C.~S., Farquhar, G., Nardelli, N., Rudner,
  T.~G., Hung, C.-M., Torr, P.~H., Foerster, J., and Whiteson, S.
\newblock The starcraft multi-agent challenge.
\newblock \emph{arXiv preprint arXiv:1902.04043}, 2019.

\bibitem[Son et~al.(2019)Son, Kim, Kang, Hostallero, and Yi]{son2019qtran}
Son, K., Kim, D., Kang, W.~J., Hostallero, D.~E., and Yi, Y.
\newblock Qtran: Learning to factorize with transformation for cooperative
  multi-agent reinforcement learning.
\newblock In \emph{International conference on machine learning}, pp.\
  5887--5896. PMLR, 2019.

\bibitem[Su et~al.(2024)Su, Zhou, Jiang, Gan, Wang, and Lu]{su2024multi}
Su, K., Zhou, S., Jiang, J., Gan, C., Wang, X., and Lu, Z.
\newblock Multi-agent alternate q-learning.
\newblock In \emph{Proceedings of the 23rd International Conference on
  Autonomous Agents and Multiagent Systems}, pp.\  1791--1799, 2024.

\bibitem[Sunehag et~al.(2017)Sunehag, Lever, Gruslys, Czarnecki, Zambaldi,
  Jaderberg, Lanctot, Sonnerat, Leibo, Tuyls, et~al.]{sunehag2017value}
Sunehag, P., Lever, G., Gruslys, A., Czarnecki, W.~M., Zambaldi, V., Jaderberg,
  M., Lanctot, M., Sonnerat, N., Leibo, J.~Z., Tuyls, K., et~al.
\newblock Value-decomposition networks for cooperative multi-agent learning.
\newblock \emph{arXiv preprint arXiv:1706.05296}, 2017.

\bibitem[Tan(1993)]{tan1993multi}
Tan, M.
\newblock Multi-agent reinforcement learning: Independent vs. cooperative
  agents.
\newblock In \emph{Proceedings of the tenth international conference on machine
  learning}, pp.\  330--337, 1993.

\bibitem[Wang et~al.(2020{\natexlab{a}})Wang, Ren, Liu, Yu, and
  Zhang]{wang2020qplex}
Wang, J., Ren, Z., Liu, T., Yu, Y., and Zhang, C.
\newblock Qplex: Duplex dueling multi-agent q-learning.
\newblock \emph{arXiv preprint arXiv:2008.01062}, 2020{\natexlab{a}}.

\bibitem[Wang et~al.(2020{\natexlab{b}})Wang, Ke, Qiao, and
  Chai]{wang2020large}
Wang, X., Ke, L., Qiao, Z., and Chai, X.
\newblock Large-scale traffic signal control using a novel multiagent
  reinforcement learning.
\newblock \emph{IEEE transactions on cybernetics}, 51\penalty0 (1):\penalty0
  174--187, 2020{\natexlab{b}}.

\bibitem[Wei \& Luke(2016)Wei and Luke]{wei2016lenient}
Wei, E. and Luke, S.
\newblock Lenient learning in independent-learner stochastic cooperative games.
\newblock \emph{Journal of Machine Learning Research}, 17\penalty0
  (84):\penalty0 1--42, 2016.

\bibitem[Yu et~al.(2022)Yu, Velu, Vinitsky, Gao, Wang, Bayen, and
  Wu]{yu2022surprising}
Yu, C., Velu, A., Vinitsky, E., Gao, J., Wang, Y., Bayen, A., and Wu, Y.
\newblock The surprising effectiveness of ppo in cooperative multi-agent games.
\newblock \emph{Advances in Neural Information Processing Systems},
  35:\penalty0 24611--24624, 2022.

\bibitem[Zhong et~al.(2024)Zhong, Kuba, Feng, Hu, Ji, and
  Yang]{zhong2024heterogeneous}
Zhong, Y., Kuba, J.~G., Feng, X., Hu, S., Ji, J., and Yang, Y.
\newblock Heterogeneous-agent reinforcement learning.
\newblock \emph{Journal of Machine Learning Research}, 25\penalty0
  (32):\penalty0 1--67, 2024.

\bibitem[Zhou et~al.(2021)Zhou, Luo, Villella, Yang, Rusu, Miao, Zhang, Alban,
  Fadakar, Chen, et~al.]{zhou2021smarts}
Zhou, M., Luo, J., Villella, J., Yang, Y., Rusu, D., Miao, J., Zhang, W.,
  Alban, M., Fadakar, I., Chen, Z., et~al.
\newblock Smarts: An open-source scalable multi-agent rl training school for
  autonomous driving.
\newblock In \emph{Conference on robot learning}, pp.\  264--285. PMLR, 2021.

\end{thebibliography}
\bibliographystyle{icml2026}

\newpage
\appendix
\onecolumn

\section{The Distinction Clarification}
\label{sec:the_distinction_clarification}
For each agent $i$, the decentralized value function $Q^{i}(s_{t}, a_{t}^{i})$ can be regarded as a projection of the true joint action value function $Q(s_{t},a_{t}^{i},a_{t}^{-i})$. Specifically, IQL adheres to an average-based projection defined as follows:
\begin{equation}
	\begin{aligned}
		Q^{i,\boldsymbol{\pi}}(s_{t},a_{t}^{i})=\sum\nolimits_{a_{t}^{-i}}\pi^{-i}(a_{t}^{-i}|s_{t})Q^{\boldsymbol{\pi}}(s_{t},a_{t}^{i},a_{t}^{-i}),
	\end{aligned}
\end{equation}
where $Q^{\boldsymbol{\pi}}(s_{t},a_{t}^{i},a_{t}^{-i})$ represents the joint action value function given a joint policy $\boldsymbol{\pi}=(\pi^{i}, \pi^{-i})$. It is obvious that the average-based projection is easily affected by other agents' sub-optimal actions and suffers from the relative overgeneralization. In contrast, the maximum-based (optimistic) projection is defined below:
\begin{equation}
	\begin{aligned}
		\label{eq:optimistic_projection}
		Q^{i,\operatorname{opt}}(s_{t},a_{t}^{i})=\max\nolimits_{a_{t}^{-i}}Q^{*}(s_{t},a_{t}^{i},a_{t}^{-i}),
	\end{aligned}
\end{equation}
where $Q^{*}(s_{t},a_{t}^{i},a_{t}^{-i})$ represents the joint action value function of an optimal joint policy $\boldsymbol{\pi}^{*}$. This optimistic projection assumes that other agents $-i$ always select their local cooperative actions, thereby eliminating the impact of other agents' non-cooperation (sub-optimal or exploratory action selections). As a result, each agent $i$ can identify and select its local cooperative action based on $Q^{i,\operatorname{opt}}(s_{t},a_{t}^{i})$, leading to the optimal joint policy.

\textbf{Distributed Q-learning.} Distributed Q-learning introduces an optimistic value function update for $Q^{i}(s_{t}, a_{t}^{i})$ to directly approximate $Q^{i,\operatorname{opt}}(s_{t},a_{t}^{i})$. Specifically, $Q^{i}(s_{t}, a_{t}^{i})$ is updated as follows:
\begin{equation}
	\label{eq:distributed_q_update}
	Q^{i}(s_{t}, a^{i}_{t})\leftarrow \left\{
	\begin{aligned}
		& Q^{i}(s_{t}, a^{i}_{t}) + \delta^{i}_{t} \quad \operatorname{if} \delta^{i}_{t} \geq 0 \\ 
		& Q^{i}(s_{t}, a^{i}_{t}) \quad \quad \quad \operatorname{else}
	\end{aligned}  
	\right.,
\end{equation}
where $Q^{i}(s_{t}, a_{t}^{i})$ is solely updated when the temporal difference error (TD error) $\delta_{t}^{i}$ is positive. $\delta_{t}^{i}$ is defined as follows:
\begin{equation}
	\begin{aligned}
		\delta_{t}^{i}=\underbrace{r_{t}+\gamma\max_{a_{t+1}^{i}}Q^{i}(s_{t+1}, a_{t+1}^{i})}_{\operatorname{learning \ target}} - Q^{i}(s_{t}, a_{t}^{i}).
	\end{aligned}
	\label{eq_2}
\end{equation}

\textbf{Hysteretic Q-learning.} However, due to the high optimism, Distributed Q-learning is vulnerable to stochasticity. To address this issue, Hysteretic Q-learning updates $Q^{i}(s_{t}, a_{t}^{i})$ using two learning rates for positive and negative TD errors, respectively. Specifically, Hysteretic Q-learning updates $Q^{i}(s_{t}, a_{t}^{i})$ as follows:
\begin{equation}
	\label{eq:hysteretic_q_update}
	\begin{aligned}
		Q^{i}(s_{t}, a^{i}_{t})\leftarrow \left\{
		\begin{aligned}
			& Q^{i}(s_{t}, a^{i}_{t}) + \delta^{i}_{t} \quad \operatorname{if} \delta^{i}_{t} \geq 0 \\ 
			& Q^{i}(s_{t}, a^{i}_{t}) + \beta\delta^{i}_{t} \quad \operatorname{else}
		\end{aligned}  
		\right.,
	\end{aligned}
\end{equation}
where $\beta<1$ is a complement factor to control the contribution of negative $\delta_{t}^{i}$ to the update of $Q^{i}(s_{t}, a_{t}^{i})$. 

Although Distributed Q-learning theoretically demonstrates that $Q^{i}(s_{t},a_{t}^{i})$ with the optimistic value update converges to the optimal joint policy, the entire update process (both Eq.~(\ref{eq:distributed_q_update}) and Eq.~(\ref{eq:hysteretic_q_update})) solely relies on the decentralized value function $Q^{i}(s_{t},a_{t}^{i})$. When facing with complex cooperative tasks where multiple agents strongly influence each other, such fully decentralized value function update often leads to instability and poor convergence. Furthermore, such optimistic value function update with function approximators (particularly the deep neural networks) is susceptible to the overestimation issue, resulting in sub-optimal solutions. The poor performance of Hysteretic Q-learning empirically validates this insight.

\textbf{Dynamics-Aware Context (DAC).} In contrast, DAC decomposes the approximation of $Q^{i,\operatorname{opt}}(s_{t}, a_{t}^{i})$ into two sub-processes. The first is that DAC learns a context-based value function $Q^{i}(s_{t}, a_{t}^{i}, c_{t}^{i})$. The context $c_{t}^{i}$ that models the agent $i$'s current local dynamics distribution implicitly represents the current joint policy $\pi^{-i}$ of other agents $-i$. As a result, $Q^{i}(s_{t}, a_{t}^{i}, c_{t}^{i})$ approximates the true joint action value function $Q(s_{t}, a_{t}^{i}, a_{t}^{-i})$, as empirically demonstrated by visualizations in the matrix game. Furthermore, the contexts enable stationary value function update of $Q^{i}(s_{t}, a_{t}^{i}, c_{t}^{i})$. According to the Bellman update, $Q^{i}(s_{t}, a_{t}^{i}, c_{t}^{i})$ finally converges to the optimal joint action value function $Q^{*}(s_{t}, a_{t}^{i}, a_{t}^{-i})$.

The second is that we further derive the optimistic marginal value $\phi^{i}(s_{t}, a_{t}^{i})$ according to $\phi^{i}(s_{t}, a_{t}^{i})=\max_{c\in\mathcal{C}}Q^{i}(s_{t}, a_{t}^{i}, c)$. As $Q^{i}(s_{t}, a_{t}^{i}, c_{t}^{i})$ converges to $Q^{*}(s_{t},a_{t}^{i},a_{t}^{-i})$ following a stationary value update, $\phi^{i}(s_{t}, a_{t}^{i})$ approaches $Q^{i, \operatorname{opt}}(s_{t}, a_{t}^{i})$.

Based on these two sub-processes, we ensure the stationary approximation of $Q^{i, \operatorname{opt}}(s_{t}, a_{t}^{i})$ by the optimistic marginal value $\phi^{i}(s_{t}, a_{t}^{i})=\max_{c\in\mathcal{C}}Q^{i}(s_{t}, a_{t}^{i}, c)$. In comparison to Distributed Q-learning and Hysteretic Q-learning, the context-based value function of DAC is free from the non-stationarity and overestimation issues. The superior performance of DAC across various cooperative tasks further validates its effectiveness in enhancing fully decentralized cooperative policy learning.

\section{Theoretical Derivation}
\label{sec:theoretical_derivation}
For modeling the task dynamics distribution of $\tau_{t-k+1:t}^{i}$, we assume that this distribution can be represented by a latent variable $c_{t}^{i}$, and the underlying mapping from the trajectory segment to the variable adheres to an unknown probability distribution $p^{i}(c_{t}^{i}|\tau_{t-k+1:t}^{i})$. We learn a distribution $q^{i}(c_{t}^{i}|\tau_{t-k+1:t}^{i})$ to approximate it, and optimize this approximation by minimizing the KL-divergence between them:
\begin{equation}
	\label{eq:complete_vae_inference}
	\begin{aligned}
		D_{\operatorname{KL}}&(q^{i}(c_{t}^{i}|\tau_{t-k+1:t}^{i})||p^{i}(c_{t}^{i}|\tau_{t-k+1:t}^{i})) \\
		&=\mathbb{E}_{q^{i}(c_{t}^{i}|\tau_{t-k+1:t}^{i})}\log\frac{q^{i}(c_{t}^{i}|\tau_{t-k+1:t}^{i})}{p^{i}(c_{t}^{i}|\tau_{t-k+1:t}^{i})} \\
		&=\mathbb{E}_{q^{i}(c_{t}^{i}|\tau_{t-k+1:t}^{i})}\log q^{i}(c_{t}^{i}|\tau_{t-k+1:t}^{i}) - \mathbb{E}_{q^{i}(c_{t}^{i}|\tau_{t-k+1:t}^{i})}\log p^{i}(c_{t}^{i}|\tau_{t-k+1:t}^{i}) \\
		&=\mathbb{E}_{q^{i}(c_{t}^{i}|\tau_{t-k+1:t}^{i})}\log q^{i}(c_{t}^{i}|\tau_{t-k+1:t}^{i}) - \mathbb{E}_{q^{i}(c_{t}^{i}|\tau_{t-k+1:t}^{i})}\log \frac{p^{i}(c_{t}^{i}, \tau_{t-k+1:t}^{i})}{p^{i}(\tau_{t-k+1:t}^{i})} \\
		&=\mathbb{E}_{q^{i}(c_{t}^{i}|\tau_{t-k+1:t}^{i})}\log q^{i}(c_{t}^{i}|\tau_{t-k+1:t}^{i}) - \mathbb{E}_{q^{i}(c_{t}^{i}|\tau_{t-k+1:t}^{i})}\log \frac{p^{i}(\tau_{t-k+1:t}^{i}|c_{t}^{i})p^{i}(c_{t}^{i})}{p^{i}(\tau_{t-k+1:t}^{i})} \\
		&=\mathbb{E}_{q^{i}(c_{t}^{i}|\tau_{t-k+1:t}^{i})}\log \frac{q^{i}(c_{t}^{i}|\tau_{t-k+1:t}^{i})}{p^{i}(c_{t}^{i})}- \mathbb{E}_{q^{i}(c_{t}^{i}|\tau_{t-k+1:t}^{i})}\log p^{i}(\tau_{t-k+1:t}^{i}|c_{t}^{i}) + \mathbb{E}_{q^{i}(c_{t}^{i}|\tau_{t-k+1:t}^{i})}\log p^{i}(\tau_{t-k+1:t}^{i}) \\
		&=\log p^{i}(\tau_{t-k+1:t}^{i})+D_{\operatorname{KL}}(q^{i}(c_{t}^{i}|\tau_{t-k+1:t}^{i})||p^{i}(c_{t}^{i}))-\mathbb{E}_{q^{i}(c_{t}^{i}|\tau_{t-k+1:t}^{i})}\log p^{i}(\tau_{t-k+1:t}^{i}|c_{t}^{i}),
	\end{aligned}
\end{equation}
where $p^{i}(c_{t}^{i})$ denotes the true prior distribution of the latent variable, and $\log p^{i}(\tau_{t-k+1:t}^{i})$ is the evidence that can be regarded as a constant. Based on Eq.~(\ref{eq:complete_vae_inference}), minimizing $D_{\operatorname{KL}}(q^{i}(c_{t}^{i}|\tau_{t-k+1:t}^{i})||p^{i}(c_{t}^{i}|\tau_{t-k+1:t}^{i}))$ can be achieved by the equation below:
\begin{equation}
	\label{eq:vae_optimize_appendix}
	\begin{aligned}
		\max \mathbb{E}_{q^{i}(c_{t}^{i}|\tau_{t-k+1:t}^{i})}\log p^{i}(\tau_{t-k+1:t}^{i}|c_{t}^{i})-D_{\operatorname{KL}}(q^{i}(c_{t}^{i}|\tau_{t-k+1:t}^{i})||p^{i}(c_{t}^{i})).
	\end{aligned}
\end{equation}

\begin{figure*}[t]
	\centering
	\includegraphics[width=1.0\linewidth]{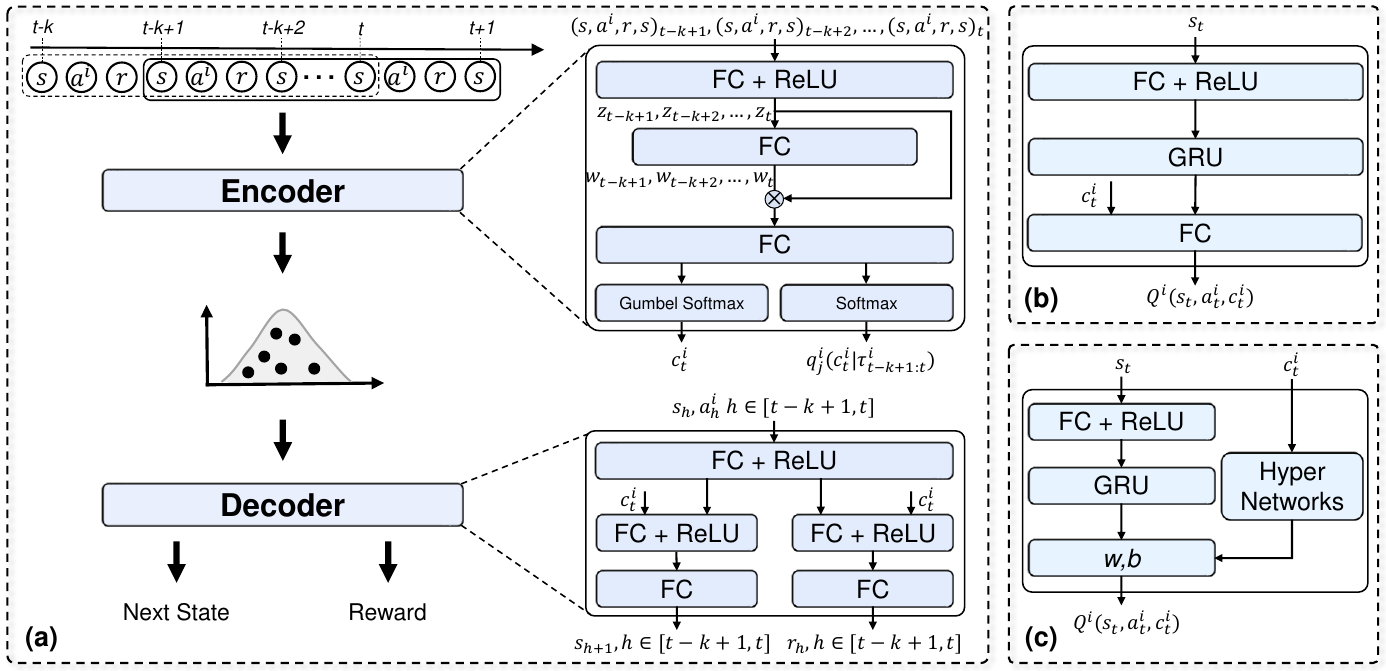}
	\caption{The architecture of DAC. (a) The VAE-like network which contains the encoder and decoder modules. (b) The Normal-Net implementation of the context-based value function $Q^{i}(s_{t}, a_{t}^{i}, c_{t}^{i})$. (c) The Hyper-Net implementation of the context-based value function $Q^{i}(s_{t}, a_{t}^{i}, c_{t}^{i})$. Note that DAC maintains a sliding window to hold the latest $k$ transitions.}
	\label{fig:dac_architecture_appendix}
\end{figure*}

\section{Algorithmic Details}
\label{sec:algorithmic_details}
\subsection{Implementation}
The core of DAC lies in learning the context-based value function $Q^{i}(s_{t}, a_{t}^{i}, c_{t}^{i})$. As shown in Fig.~\ref{fig:dac_architecture_appendix}, we provide two kinds of implementations. (1) Normal-Net. This architecture contains a hidden layer (64 units, ReLU activation), a GRU module, and a linear layer (64 units). The final linear layer takes the hidden states and contexts as inputs, and outputs the Q values of all local actions. (2) Hyper-Net. It is comprised by a hidden layer (64 units, ReLU activation), a GRU module, and a linear layer (64 units). The weights and biases of the final linear layer are generated by two separate hyper-networks, which take the context as inputs. The other implementation details are the same as standard IQL in PyMARL~\cite{samvelyan2019starcraft}. 

\subsection{Pseudocode}
As presented in Algorithm~\ref{alg:DAC}, the learning procedure of DAC is as follows. We begin by initializing the context-based value function, the encoder, and the decoder for each agent $i\in N$. During each episode, each agent $i$ selects its local action based on $q^{i}(s_{t}, a_{t}^{i})$ during exploration and $\phi^{i}(s_{t}, a_{t}^{i})$ during testing. After each episode terminates, the local episode data is stored into each agent $i$'s local replay buffer $\mathcal{D}^{i}$. During training, we sample batches of episodes from $\mathcal{D}^{i}$, and constructs the set of all trajectory segments $\tau_{t-k+1:t}^{i}$ for each episodic trajectory $\tau^{i}$. Subsequently, we shape the corresponding contexts $c_{t}^{i}$ using the encoder $q^{i}(c_{t}^{i}|\tau_{t-k+1:t}^{i})$. Finally, we calculate the value loss $\mathcal{L}_{\operatorname{C}}$, the KL divergence loss $\mathcal{L}_{\operatorname{KL}}$, and the reconstruction loss $\mathcal{L}_{\operatorname{REC}}$, and update all components. The entire procedure continues until the maximum training episode is reached.

\begin{algorithm}[t]
	\caption{Dynamics-Aware Context (DAC)}
	\label{alg:DAC}
	Initialize necessary hyper-parameters \\
	\For{agent $i\in N$}{
		Initialize parameters $\theta^{i}, \omega_{\operatorname{e}}^{i}, \omega_{\operatorname{d}}^{i}$ of the context-based value function, the encoder, and the decoder \\
	}
	\If{the maximum training episode is not reached}{
		\For{Each episode}{
			\ForEach{time step $t$}{
				\ForEach{agent $i\in N$}{
					\If{train}{
						Select its local action $a_{t}^{i}$ based on $q^{i}(s_{t}, a_{t}^{i})$ (Eq.~\ref{eq:context_sampling}) \\
					}
					\If{test}{
						Select its local action $a_{t}^{i}$ based on $\phi^{i}(s_{t}, a_{t}^{i})$ (Eq.~\ref{eq:optimistic_marginal_value}) \\
					}
				}
			}
		}
		Store each agent $i$'s episode into its individual replay buffer $\mathcal{D}^{i}$ \\
		\If{train}{
			\ForEach{agent $i\in N$}{
				Sample batches of episodes from $\mathcal{D}^{i}$ \\
				Construct the set of trajectory segments $\tau_{t-k+1:t}^{i}$ (\emph{i.e.,} the sliding window) for each episodic trajectory $\tau^{i}$ \\ 
				Construct contexts $c_{t}^{i}$ based on the encoder (\emph{i.e.,} the mapping distribution) $q^{i}(c_{t}^{i}|\tau_{t-k+1:t}^{i})$ \\
				Calculate $\mathcal{L}_{\operatorname{C}}(\theta^{i}), \mathcal{L}_{\operatorname{KL}}(\omega_{\operatorname{e}}^{i}), \mathcal{L}_{\operatorname{REC}}(\omega_{\operatorname{e}}^{i}, \omega_{\operatorname{d}}^{i})$ according to Eq.~(\ref{eq:context_value_function}), Eq.~(\ref{eq:kl_loss_function}) and Eq.~(\ref{eq:vae_loss_function}) \\
				Update all components with gradient descent \\
			}
		}
	}
\end{algorithm}

\section{Experimental Details}
\label{sec:experimental_details}
\subsection{Benchmarks}
\textbf{Matrix Game.} Matrix game is a simple two-agent cooperative stage game where two agents must select the optimal joint action $(a^{1}, a^{1})$ to receive the best reward $8$. In addition to the optimal joint action $(a^{1}, a^{1})$, there are two sub-optimal joint actions $(a^{2}, a^{2})$ and $(a^{3}, a^{3})$ that lead to $6$ rewards. For each agent, when the other agent uniformly selects its local actions, the sub-optimal local actions $a^{2}$ and $a^{3}$ may be preferred over the optimal ones $a^{1}$, leading to the relative overgeneralization.

\textbf{Predator and Prey.} We adopt the modified predator-prey~\cite{son2019qtran} to further validate DAC's effectiveness. This task involves a 5$\times$5 grid world wherein 2 predators must coordinate to capture 1 moving prey. When the sole prey is within the cardinal direction of both predators, it is ``captured'' and then regenerated at random positions. The observation of each predator agent includes its own coordinates, agent ID, and the relative coordinates of the prey, and the action set contains moving left, right, up, down, and stop. A team reward of +1 is emitted when two predator agents simultaneously capture the prey, otherwise a penalty of -$P$ is provided for sole pursuit. An entire episode proceeds over 100 time steps, and we test four settings $P=0.0, 0.3, 0.5, 1.0$ where a larger value of $P$ induces more significant relative overgeneralization challenge.

\textbf{SMAC.} The StarCraft Multi-Agent Challenge (SMAC) serves as a widely used benchmark in which a set of challenging maps is provided. These maps require cooperative MARL algorithms to make decentralized control for allied units against build-in AI enemies and achieve high win rates. In this paper, we choose seven maps: 5m\_vs\_6m, 8m\_vs\_9m, 10m\_vs\_11m, 3s\_vs\_4z, 3s\_vs\_5z, 2s\_vs\_1sc, and 3s5z to evaluate our algorithm. Details regarding these maps can be found in Tab.~\ref{table:maps}.

\begin{table}[t]
	\caption{Descriptions of maps used in this paper.}
	\label{table:maps}
	\centering
	\resizebox{0.65\linewidth}{!}{
		\begin{tabular}{c|c|c|c}
			\toprule
			Name	& Ally Units & Enemy Units & Type \\
			\midrule
			2s\_vs\_1sc & 2 Stalkers & 1 Spine Crawler & Asymmetric \& Homogeneous \\
			\hline
			\multirow{2}{*}{3s5z}	& 3 Stalkers, & 3 Stalkers, & \multirow{2}{*}{Symmetric \& Heterogeneous} \\
			& 5 Zealots & 5 Zealots & \\
			\hline
			3s\_vs\_4z & 3 Stalkers & 4 Zealots & Asymmetric \& Homogeneous \\
			\hline
			3s\_vs\_5z & 3 Stalkers & 5 Zealots & Asymmetric \& Homogeneous \\
			\hline
			5m\_vs\_6m & 5 Marines & 6 Marines & Asymmetric \& Homogeneous \\
			\hline
			8m\_vs\_9m & 8 Marines & 9 Marines & Asymmetric \& Homogeneous \\
			\hline
			10m\_vs\_11m & 10 Marines & 11 Marines & Asymmetric \& Homogeneous \\
			\bottomrule
	\end{tabular}}
\end{table}

\subsection{Baselines}
In this paper, we compare DAC with several baselines, namely IQL, Hysteretic Q-learning, and I2Q, as presented in Tab.~\ref{table:algorithms}. Below is a brief introduction to each algorithm.

\textbf{IQL.} IQL learns a decentralized value function $Q^{i}(s_{t},a_{t}^{i})$ for each agent $i$, and updates it following the standard Q-learning paradigm. However, since IQL agent ignores other agents' actions and policies, it suffers from non-stationarity during value function updates. What's more, the $Q^{i}(s_{t}, a_{t}^{i})$ learned by IQL is an average-based projection of the true joint action value function, which induces relative overgeneralization issue during value estimation, as stated in Sec.~\ref{sec:mmdp}.

\textbf{Hysteretic Q-learning.} This algorithm employs an optimistic value function update (Eq.~(\ref{eq:hysteretic_q_update})) for each agent $i$'s $Q^{i}(s_{t}, a_{t}^{i})$ to directly approximate the optimistic projection $Q^{i, \operatorname{opt}}(s_{t}, a_{t}^{i})$. Such optimistic update assumes that other agents always select their cooperative actions, thereby eliminating the negative effects caused by other agents' sub-optimal or exploratory action selections and addressing the relative overgeneralization issue. However, the non-stationarity issue remains unsolvable.

\textbf{I2Q.} I2Q addresses both non-stationarity and relative overgeneralization by shaping ideal transitions, which are constructed by selecting next states with the highest QSS value. These transitions implicitly assume that other agents follow cooperative local policies. By learning based on such transitions, I2Q guides IQL agents toward optimal cooperative policies.

\begin{table}[h]
	\caption{Challenges faced by all baselines.}
	\label{table:algorithms}
	\centering
	\resizebox{0.55\columnwidth}{!}{
		\begin{tabular}{c|c|c}
			\toprule
			\multirow{1}{*}{Algorithm}	& Non-stationarity & Relative Overgeneralization \\
			\midrule
			IQL	& \Checkmark & \Checkmark \\
			\hline
			Hysteretic Q-learning	& \Checkmark & \XSolidBrush \\
			\hline
			I2Q	& \XSolidBrush & \XSolidBrush \\
			\bottomrule
		\end{tabular}
	}
\end{table}

\subsection{Experimental Setups}
\label{sec:experimental_setups}
We implement all algorithms using the PyMARL framework. For I2Q, we adopt its official source code and the recommended hyper-parameters to conduct experiments on predator and prey and SMAC maps. For the matrix game, we directly report I2Q's results from the original paper. For IQL, we employ the original implementation provided in PyMARL. For Hysteretic Q-learning, we set $\beta$=0.01 for the matrix game. For predator and prey and SMAC maps, we set $\beta$=0.01, 0.1, 0.3, 0.5, and 0.7, and run multiple seeds using $\beta$ that yields the best performance.

For our method DAC, we instantiate the context-based value function $Q^{i}(s_{t}, a_{t}^{i}, c_{t}^{i})$ using Hyper-Net architecture to evaluate its performance on the matrix game and predator and prey tasks. For the SMAC maps, we use Normal-Net to implement $Q^{i}(s_{t},a_{t}^{i},c_{t}^{i})$. Empirically, we find these configurations work well. All other settings, such as learning rate, batch size, are kept consistent across all algorithms. The specific hyper-parameter settings of DAC across all tasks are provided in Tab.~\ref{table:hyper_params}. In particular, $k$ denotes the sliding window length and $m$ is the context dimension (\emph{i.e.,} the number of discrete contexts), while $\beta$ is the scaling factor that balances the KL loss and the reconstruction loss. $T_{epsilon}$ denotes the anneal time steps of $\epsilon$ when $\epsilon$-greedy policy is used for exploration, and we decrease $\epsilon$ from $\epsilon_{max}$ to $\epsilon_{min}$ within $T_{epsilon}$ time steps. $T_{max}$ is the total number of training time steps, and $N_{max}$ denotes the size of each agent's replay buffer. $N_{batch}$ represents the size of sampled batches per training. $\alpha$ is the learning rate and $\gamma$ is the discounted factor.

\begin{figure*}[t]
	\centering
	\includegraphics[width=1.0\linewidth]{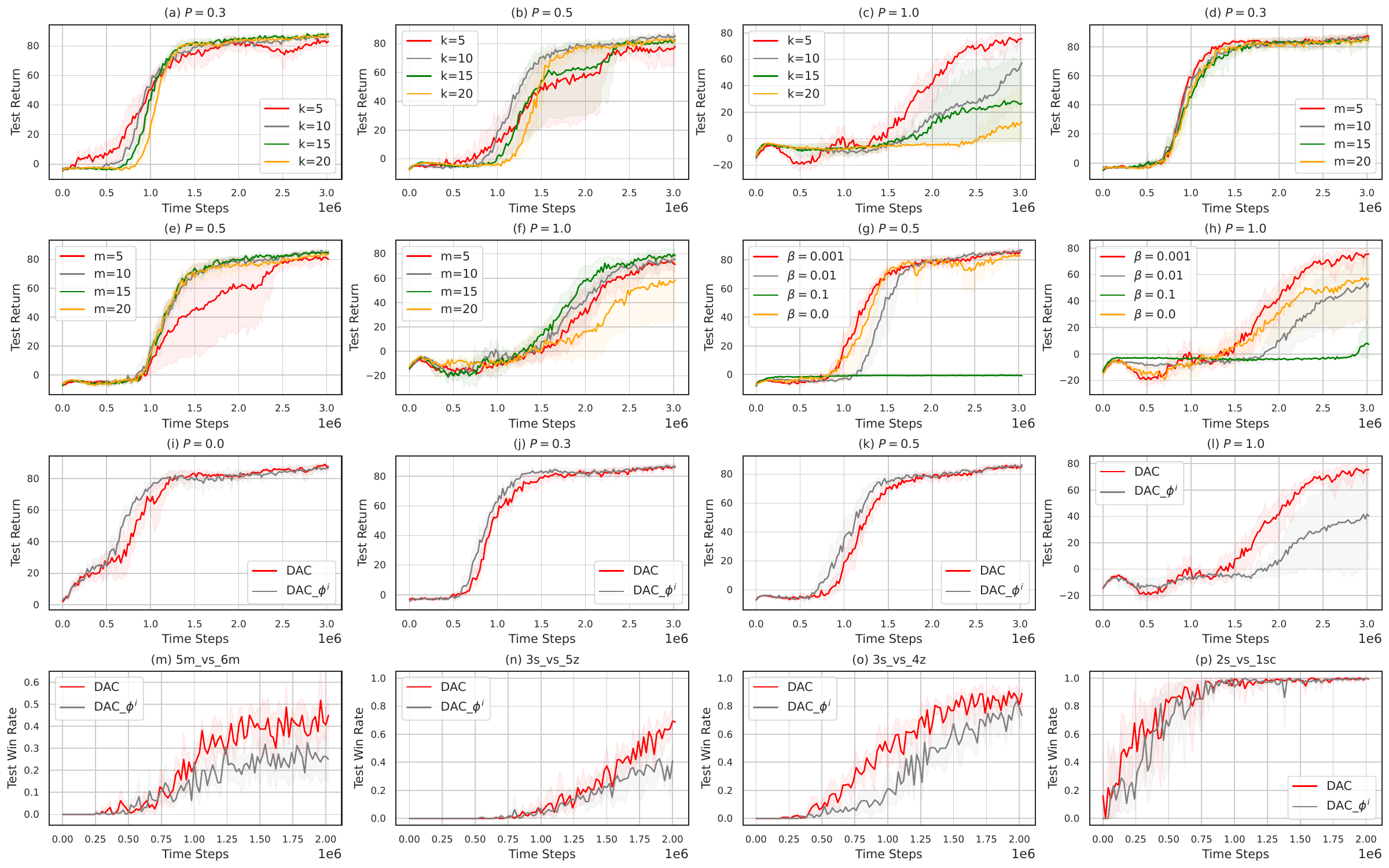}
	\caption{Ablation study regarding major components of DAC.}
	\label{fig:ablation_study}
\end{figure*}

\subsection{Ablation Study}
\label{sec:ablation_study_appendix}
To examine the effects of DAC's major components: (1) the sliding window length ($k$), (2) the number of discrete contexts ($m$), (3) the KL loss coefficient ($\beta$), and (4) the augmented exploration, we conduct a comprehensive ablation study. For components (1), (2), and (3), we evaluate $k \in \{5,10,15,20\}$, $m \in \{5,10,15,20\}$, and $\beta \in \{0.0, 0.001, 0.01, 0.1\}$. For component (4), we design a variant, namely DAC\_$\phi^{i}$, in which agents select actions solely according to $\phi^{i}(s_{t}, a_{t}^{i})$.

As depicted in Fig.~\ref{fig:ablation_study}, DAC with $m=20$ exhibits degraded performance in predator-prey with $P=1.0$. This indicates that an excessively large $m$ results in inefficient updates of the context-based value function due to the expanded context space, particularly in challenging tasks where relative overgeneralization limits effective reward signals. Conversely, a small number of discrete contexts (\emph{e.g.,} $m=5$) fails to represent diverse local task dynamics and the corresponding other agents' diverse joint policies, as evidenced by its poor performance in predator-prey with $P=0.5$.

Regarding the sliding window length $k$, a small value may lead to abrupt fluctuations in the inferred dynamics distribution, while a large value may fail to capture timely distributional changes. The superior performance of DAC with $k=10$ in comparison to $k=5, 15, 20$ in predator-prey with $P=0.3$ and $0.5$ supports this observation. However, DAC with $k=5$ achieves strong performance in predator-prey with $P=1.0$, indicating that the optimal choice of $k$ is task-dependent and should be selected empirically based on task difficulty.

For the KL loss coefficient, DAC with $\beta=0.001$ consistently outperforms $\beta=0.0$, indicating that the KL loss encourages effective utilization of all discrete contexts and accordingly results in better performance. However, overly large values (\emph{e.g.,} $\beta=0.01$ and $0.1$) degrade performance, highlighting the need to balance reconstruction and KL regularization objectives.

In addition, DAC outperforms DAC\_$\phi^{i}$ in complex tasks while achieving comparable performance on simpler tasks, which demonstrates that efficient exploration is essential for comprehensive coverage of per-agent local task dynamics and accurate estimation of context-based value functions, as well as DAC's performance.

In summary, (4) the augmented exploration is critical for achieving sufficient coverage of local task dynamics and enabling accurate context-based value estimation. In terms of (1) the sliding window length and (2) the number of discrete contexts, DAC currently relies on empirical tuning to specify the selection of $k$ and $m$. More efforts about adaptive context modeling are necessitated to reduce this burden. For (3) the KL loss, it is critical to enable sufficient usage of all possible discrete contexts during dynamics distribution modeling, and we empirically find that $\beta=0.001$ works well across various tasks.

\begin{table*}[t]
	\caption{Hyper-parameters of DAC across all tasks.}
	\label{table:hyper_params}
	\centering
	\resizebox{1.0\linewidth}{!}{
		\begin{tabular}{c|c|c|c|c|c|c|c|c|c|c|c|c}
			\toprule
			\multirow{2}{*}{Tasks} & \multirow{2}{*}{$k$} & \multirow{2}{*}{$m$} & \multirow{2}{*}{$\beta$} & \multirow{2}{*}{$T_{epsilon}$} & \multirow{2}{*}{$T_{max}$} & \multirow{2}{*}{$N_{max}$} & \multirow{2}{*}{$N_{batch}$} & \multirow{2}{*}{Optimizer} & \multirow{2}{*}{$\alpha$} & \multirow{2}{*}{$\gamma$} & \multirow{2}{*}{$\epsilon_{max}$} & \multirow{2}{*}{$\epsilon_{min}$} \\
			~ & ~ & ~ & ~ & ~ & ~ & ~ & ~ & ~ & ~ & ~ & ~ \\
			\midrule
			matrix game	& 1 & 4 & \multirow{12}{*}{0.001} & 50k & 50000 & \multirow{12}{*}{5000} & \multirow{12}{*}{32} & \multirow{12}{*}{RMSprop} & \multirow{12}{*}{0.0005} & \multirow{12}{*}{0.99} & \multirow{12}{*}{1.0} & \multirow{12}{*}{0.05} \\
			
			$P=0.0$ & 10 & 10 & ~ & 1500k & 3050000 & ~ & ~ & ~ & ~ & ~ & ~ & ~  \\
			
			$P=0.3$ & 10 & 10 & ~ & 1500k & 3050000 & ~ & ~ & ~ & ~ & ~ & ~ & ~  \\
			
			$P=0.5$ & 10 & 10 & ~ & 1500k & 3050000 & ~ & ~ & ~ & ~ & ~ & ~ & ~  \\
			
			$P=1.0$ & 5 & 10 & ~ & 1500k & 3050000 & ~ & ~ & ~ & ~ & ~ & ~ & ~  \\
			
			2s\_vs\_1sc	& 10 & 10 & ~ & 50k & 2050000 & ~ & ~ & ~ & ~ & ~ & ~ & ~  \\
			
			3s5z	& 10 & 5 & ~ & 50k & 2050000 & ~ & ~ & ~ & ~ & ~ & ~ & ~   \\
			
			3s\_vs\_4z	& 15 & 20 & ~ & 50k & 2050000 & ~ & ~ & ~ & ~ & ~ & ~  & ~  \\
			
			3s\_vs\_5z	& 15 & 20 & ~ & 50k & 2050000 & ~ & ~ & ~ & ~ & ~ & ~ & ~   \\
			
			5m\_vs\_6m	& 15 & 20 & ~ & 50k & 2050000 & ~ & ~ & ~ & ~ & ~ & ~ & ~   \\
			
			8m\_vs\_9m	& 5 & 10 & ~ & 50k & 2050000 & ~ & ~ & ~ & ~ & ~ & ~ & ~   \\
			
			10m\_vs\_11m & 15 & 10 & ~ & 50k & 2050000 & ~ & ~ & ~ & ~ & ~ & ~ & ~   \\
			\bottomrule
		\end{tabular}
	}
\end{table*}

\subsection{Computational Cost}
\label{sec:computational_cost}
We run all experiments under five different random seeds, and report the mean and standard deviation across runs in all figures. The experiments are carried out on a server equipped with an AMD EPYC 7542 32-Core Processor CPU, 504GB RAM, and 8 NVIDIA GeForce RTX 4090 D GPUs.

\end{document}